\documentclass[pdflatex,sn-basic,Numbered,iicol]{sn-jnl}

\usepackage{graphicx}%
\usepackage{multirow}%
\usepackage{amsmath,amssymb,amsfonts}%
\usepackage{amsthm}%
\usepackage{mathrsfs}%
\usepackage[title]{appendix}%
\usepackage{xcolor}%
\usepackage{textcomp}%
\usepackage{manyfoot}%
\usepackage{booktabs}%
\usepackage{tabularx}
\usepackage{subfig}%
\usepackage{siunitx}
\usepackage{algorithm}%
\usepackage{algorithmicx}%
\usepackage{algpseudocode}%
\usepackage{listings}%

\theoremstyle{thmstyleone}%

\theoremstyle{thmstyletwo}%

\theoremstyle{thmstylethree}%

\begin{document}

\title[Edge-AI Condition Monitoring in Dairy Processing]{On Utilization of Edge-AI for condition monitoring towards predictive maintenance: A use case in dairy processing using explainable AI}

\author*[1]{\fnm{Rahul Umesh} \sur{Mhapsekar}}\email{rahul.mhapsekar@setu.ie}

\author[2]{\fnm{Norah} \sur{O'Shea}}\email{norah.oshea@teagasc.ie}
\author[1]{\fnm{Omer} \sur{Ali}}\email{omer.ali@setu.ie}

\author[3]{\fnm{Steven} \sur{Davy}}\email{steven.davy@tudublin.ie}

\author[1]{\fnm{Lizy} \sur{Abraham}}\email{lizy.abraham@setu.ie}

\affil*[1]{\orgname{Walton Institute}, \orgaddress{\orgname{South East Technological University}, \city{Waterford}, \country{Ireland}}}

\affil[2]{\orgdiv{Food Chemistry and Technology Department}, \orgname{Teagasc Food Research Centre}, \orgaddress{\street{Moorepark}, \city{Fermoy, Cork}, \postcode{P61 C991}, \country{Ireland}}}

\affil[1]{\orgdiv{Department of Computing and Mathematics}, \orgname{South East Technological University}, \orgaddress{\city{Carlow}, \country{Ireland}}}

\affil[3]{\orgname{TU Dublin}, \orgaddress{\city{Dublin}, \country{Ireland}}}

\abstract{The Industrial Internet of Things (IIoT) and Artificial Intelligence (AI) are driving Industry 4.0, enhancing productivity, automation, and efficiency in manufacturing systems. This study develops an Edge-AI-based condition monitoring framework, as a step towards predictive maintenance, for a rotary valve in a dairy processing pilot plant, using vibration data collected via a low-power IIoT sensor node and processed on an edge device. We first apply frequency-domain (FFT) and correlation-based analysis to identify the most informative sensor axes and reduce data dimensionality for on-edge deployment. We then compare four machine learning techniques for anomaly detection -- K-Means, DBSCAN, Isolation Forest, and Support Vector Machines -- evaluating each not only for accuracy but for memory, GPU, and energy consumption, since these constraints are critical for deployment on resource-limited edge hardware. To address the complexity and lack of transparency of these models, we further apply Explainable AI (XAI) techniques, including SHAP and Explainable Boosting Machines (EBM), to interpret model predictions, identify the most influential vibration features, and build trust in automated maintenance decisions. Results show that the supervised models reproduce a rule-based reference label derived from the RMS vibration magnitude with high fidelity (EBM 100\%, 95\% CI 98.6--100\%; linear SVM 99.61\%), and that among these models the EBM requires the least memory, energy and runtime on the edge device. These findings suggest that XAI can do more than improve interpretability: by identifying the features that drive the predictions, it can guide the design of lightweight, low-power ML models suitable for near-real-time condition monitoring in resource-constrained industrial environments.}


\maketitle

\section{Introduction}\label{sec1}
The dairy processing industry transforms raw milk from the farm into finished consumer products for the retail industry, including liquid milk, cream, yogurt, butter, ice cream, and cheese, and a range of food ingredients available for further processing, mostly in a dried powder format.  Dairy processing unit operations, including plate heat exchangers, separators, membrane filtration systems, evaporators, and spray dryers, rely on pumps, motors, and valves for efficient operation, all of which require routine maintenance \cite{b1}. These machines consist of multiple mechanical and electrical components that require routine maintenance to ensure efficient processing. The adoption of the Industrial Internet of Things (IIoT) and Artificial Intelligence (AI) in the industry is driving Industry 4.0, a new industrial revolution to enhance productivity, automation, and efficiency in manufacturing systems.

Within the IIoT framework, conventional data gathering typically involves sending extensive sensor data to the cloud for processing and analysis. This traditional cloud-centric strategy faces major challenges, including high latency, limited bandwidth, and increased energy use from continuously transmitting large sensor datasets. To overcome these challenges, Edge Computing (EC) has become a viable option, facilitating data processing near the source, thereby cutting down on transmission expenses, boosting real-time response, and increasing overall system efficiency \cite{B2A}.

In food manufacturing, maintenance strategies include corrective maintenance, which addresses failures after they occur, and preventive maintenance, which focuses on identifying and preventing potential faults. Predictive maintenance (PdM) uses AI, edge computing, and IIoT to forecast failures by analyzing past degradation patterns and future operating conditions, aiding decision-making \cite{b2}. In the context of Industry 4.0, equipment failures are anticipated by analyzing real-time data gathered from sensors embedded in industrial machinery. This sensor-driven approach enables early detection of anomalies, allowing proactive maintenance and reducing downtime. Condition monitoring serves as an essential mechanism for executing predictive maintenance plans. It depends on the ongoing and real-time observation of physical variables to identify malfunctions. A variety of equipment breakdowns can be detected through unusual trends in vibration, acoustics, or electrical current, which are evaluated in both the time and frequency spectra \cite{b3}. In this paper we distinguish condition monitoring and anomaly detection, which identify deviations from a reference behaviour, from prognostics, which forecasts failure probability or remaining useful life (RUL) and is what distinguishes PdM from condition-based monitoring~\cite{rev_jardine2006}. The present study addresses condition monitoring and anomaly detection as building blocks towards PdM.

The effectiveness of ML-based PdM depends on data quality, sensor technology, algorithm selection, and operational strategies. High-quality, real-time IoT sensor data is essential for accurate predictions, as poor data leads to suboptimal maintenance and resource waste \cite{b4,b5}. ML algorithms including deep learning and ensemble methods can process complex datasets better than statistical models \cite{b6}. When tailored to datasets, ARIMA and Double Exponential Smoothing improve prediction accuracy \cite{b7}. According to this study \cite{ba4}, various ML-models have been deployed to perform anomaly detection, however the most commonly used models include - Neural Networks such as CNN, LSTM and MLP, Outlier Detection models such as Isolation forest, Clustering models such as K-Means clustering and DBSCAN, and Support Vector Machines (SVM). Thus, it is necessary to understand the performance of these models on our application. However, neural networks are computationally intensive and consume significant energy, making deployment on mobile devices challenging \cite{ba5}. While cloud data centers traditionally handle DNN processing, the large data transfer between end devices and the cloud leads to high latency and energy consumption \cite{ba6}. Therefore, it is crucial to develop efficient models that can operate on resource-constrained edge devices and low-cost smart sensors.

State of the art PdM heavily relies on sensor data for training and real-time optimization, requiring small, wired and wireless IoT sensors that provide a flexible, adaptable data source in dynamic industrial environments, enabling efficient data acquisition \cite{b7}. Beyond technological factors, predictive maintenance also requires effective strategies for data collection, model development, and planning \cite{b8}. Despite these benefits, industrial system complexity and high implementation costs prevent widespread adoption. A multidisciplinary approach that balances technological innovation and cost-effectiveness is needed to solve these problems \cite{b9,b10,b11,b12}.

This research evaluates traditional frequency- and time-based methods and ML techniques (K-Means, DBSCAN) for real-time anomaly detection and optimization in a dairy pilot processing facility. To reduce sensor and ML model complexity, cost, and lack of transparency, Explainable Artificial Intelligence (XAI) methods are examined. They enhance trust and accountability and enable faster responses in high-stakes domains such as manufacturing and energy by clarifying how model predictions are generated. This article presents XAI analysis using SHAP, and EBM for various compute resources in a typical industrial environment, highlighting how XAI in ML-based PdM can inform model and feature selection, thereby reducing architectural complexity and response time to enhance PdM effectiveness across industrial applications.

Maintenance of rotating equipment in dairy processing is often performed on fixed schedules. At the pilot facility where this research was conducted, the rotary valve was replaced every year regardless of its condition. Data-driven condition monitoring has been demonstrated elsewhere in the agri-food chain, for instance as an edge-based maintenance system in a soybean processing plant~\cite{rev_vermesan2022} and as explainable predictive maintenance in smart agricultural facilities~\cite{rev_kisten2024}. To the best of our knowledge, however, explainable AI has not yet been applied to condition monitoring of dairy processing equipment, nor combined with on-premise edge processing in this sector. The novelty of this work lies in this application: vibration is acquired by a low-power IIoT node on an operating rotary valve and analysed on an edge device inside the plant, so that monitoring does not depend on cloud connectivity and raw process data need not leave the site. Explainable AI is used both to make each decision transparent to maintenance staff and to reveal which features and sensor axes the models rely on, which can help to reduce redundant data and model size.

The main contributions of this paper are:
\begin{enumerate}
\item An on-premise Edge-AI condition-monitoring system for dairy processing, implemented on an operating rotary valve in a dairy pilot plant.
\item A real-world vibration dataset and frequency-domain and correlation analysis of the valve at three lifecycle stages (service-exhausted, newly commissioned, and after three months of operation).
\item A resource-aware evaluation of clustering, outlier-detection and supervised models on the edge device, reporting memory, GPU utilization, energy and runtime alongside detection performance.
\item To our knowledge, the first application of explainable AI (SHAP and EBM) to condition monitoring in dairy processing, showing that a glass-box EBM can serve as both detector and explainer with the lowest resource use among the supervised models, and that its explanations recover the decision rule and identify the dominant sensor axis.
\end{enumerate}

\section{Methodology}\label{sec2}
\begin{figure}[h]
\centering
\centerline{\includegraphics[width=0.6\columnwidth]{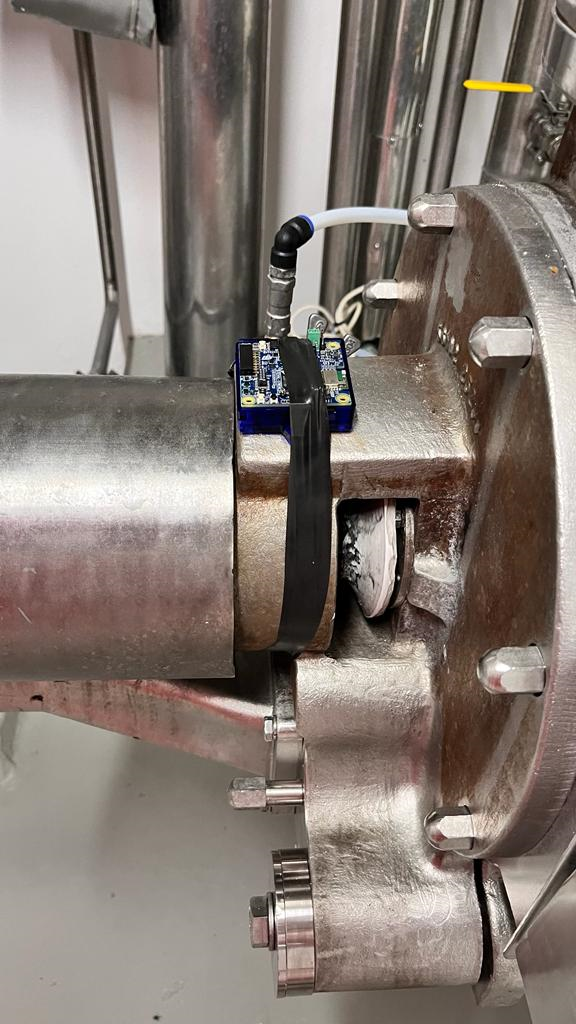}}
\caption{Data collection from the rotary valve in pilot plant}
\label{Real System}
\end{figure}

This study aims to develop an Edge-AI-based condition monitoring system, as a step towards predictive maintenance, for dairy processing equipment. Currently, no established criteria exist for characterizing the physical condition of a worn or degraded rotary valve, and the valve is replaced annually on a fixed schedule without such justification. This section outlines the methodology used to develop the proposed condition monitoring system, which applies vibrational analysis to assess the condition of the rotary valve prior to its removal. The Fig. \ref{Real System} illustrates the implementation of the proposed real-time condition monitoring system for the rotary valve at the pilot site.

The critical steps of our methodology are as follows:
\begin{figure*}[h]
\centering
\centerline{\includegraphics[width=0.8\textwidth]{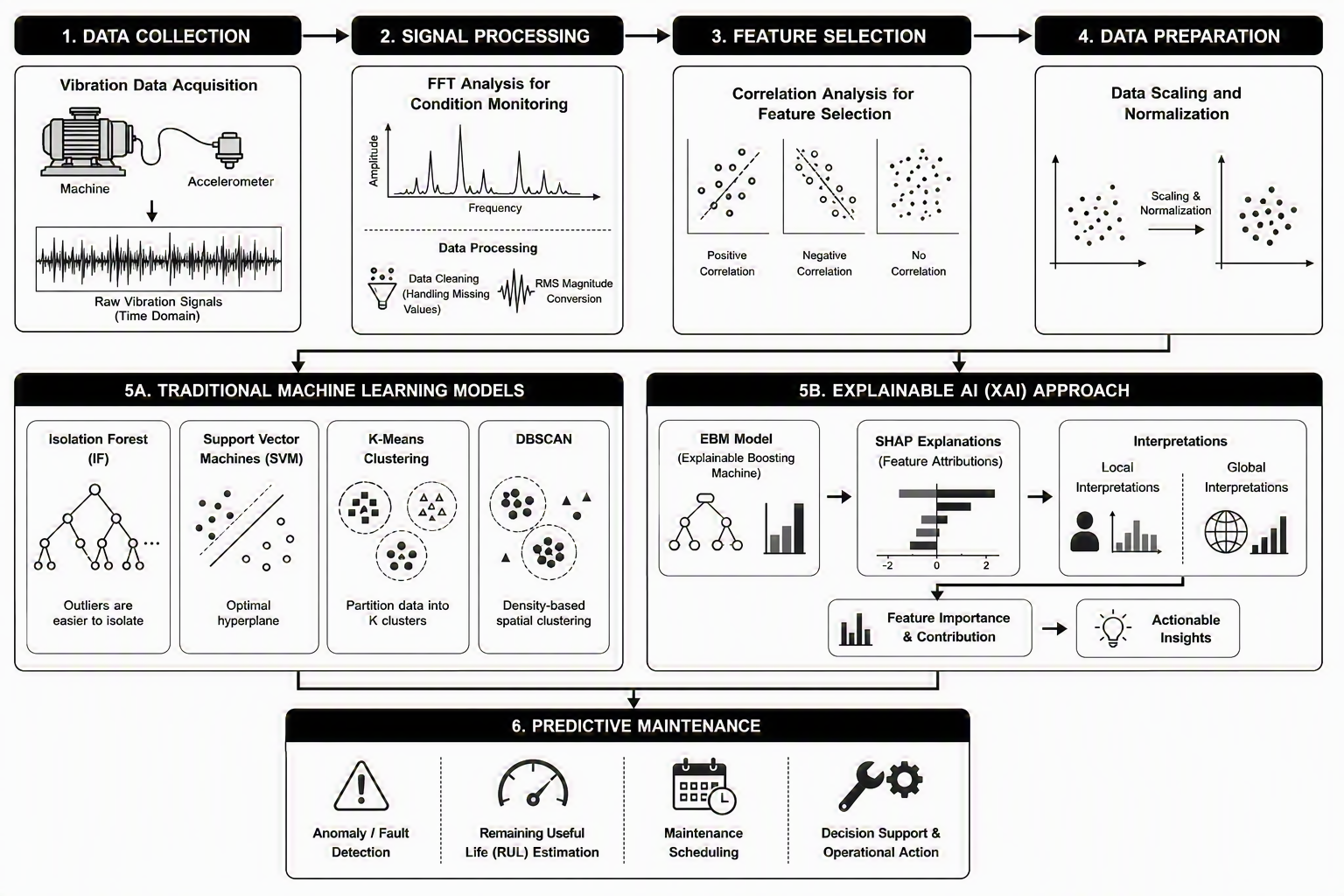}}
\caption{Predictive Maintenance framework for Rotary valve using Edge and Explainable AI for real-time monitoring}
\label{System}
\end{figure*}

\begin{figure*}[h]
\centering
\centerline{\includegraphics[width=0.95\textwidth]{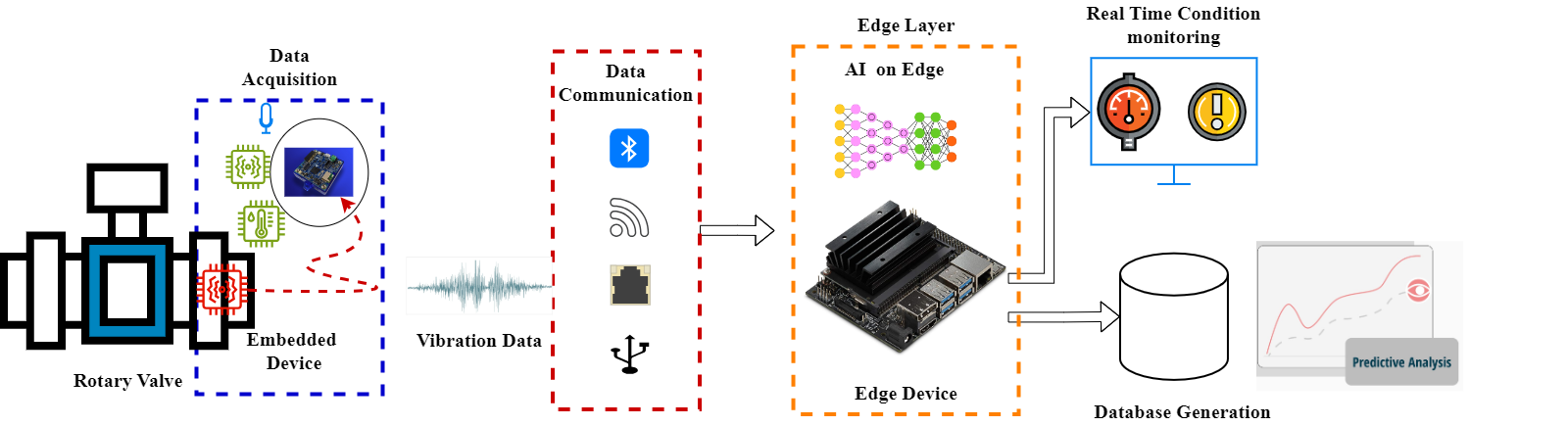}}
\caption{On-Edge condition monitoring and predictive maintenance system for Dairy processing applications}
\label{Framework System}
\end{figure*}

\begin{itemize}
    \item Data Acquisition and Preprocessing: Data is collected from a 3-axis accelerometer at three time points (service-exhausted valve, newly commissioned valve as baseline, and the same valve after 3 months). Preprocessing includes removal of timestamps, computation of the RMS magnitude, Min--Max scaling for the clustering models, and prioritizing the Y and Z axes based on correlation analysis (Section~\ref{sec:dataset}).
    \item Post-Processing and Anomaly Detection: Detected deviations from normal behavior are visualized and analyzed. Metrics such as anomaly distributions, and outlier counts are used to evaluate the condition of the valve and correlate performance with its lifecycle stage.
    \item Explainable AI Integration: Explainable AI (XAI) techniques are applied to interpret model predictions and identify the most influential features, enabling insights into valve degradation patterns and building trust in model decisions.
    \item Validation on Edge Devices: Lightweight models are executed on edge devices. Performance is evaluated using classification metrics, runtime and resource utilization (memory, GPU and energy), ensuring feasibility for on-site dairy processing applications.
\end{itemize}

\subsection{System Architecture}
\subsubsection{Experimental Setup}

The vibration data is collected using the STMicroelectronics SensorTile wireless industrial node. The device consists of a 32-bit ultra-low power microprocessor ARM cortex-M4 chip. The board also consists of various sensors such as the LSM6DSM - 3D accelerometer and 3D Gyroscope, LPS22HB - MEMS nano pressure sensor operating at 260-1260hPa, a MEMS microphone with 64dB signal-to-noise ratio for audio sensing, and LSM303AGR ultra-low power 3D Accelerometer and 3D magnetometer. The board can be used directly for data collection for condition monitoring and PdM tasks, due to the availability of all low-power sensors on a single device.

For data communication, the board includes built-in Bluetooth for wireless transmission, a USB port and SD card slot. The board can be extended for Wi-Fi communication using the STMicroelectronics Wi-Fi expansion board.  The data collected are sent to an edge processor for further analysis and processing. The edge processor is the NVIDIA Jetson Nano.
Jetson Nano is an onboard computer that consists of 4GB RAM and NVIDIA GPU for running on-edge AI applications. The models were implemented in Python with scikit-learn, InterpretML and XGBoost. The system architecture is shown in Figure \ref{Framework System}.

\subsection{Data Collection}
The data is collected from the rotary valves using the 3-axis accelerometer pre-built on the Industrial IoT node board. The placement of the board is important to get accurate axis values. The board is placed on top of the rotor for maximum acceleration from the rotor.

The data was collected in equal phases, with the first data being collected from a service exhausted valve (Old Valve), before being replaced. Second phase of data collection involved collecting data on the newly commissioned valve (baseline) which would give the ideal characteristics of a completely operational rotary valve.
The third phase included data collection on the valve after three months, to detect anomalies and observe early changes relative to the baseline; these data were not used to estimate the remaining useful life (RUL).

The real-time data is transmitted to the mobile application using Bluetooth, to provide real-time condition monitoring (surges in vibration) and also stored on the edge device for further processing.

\subsection{Dataset}\label{sec:dataset}
The study was carried out over a two-year collaboration with a pilot-scale dairy processing facility, which operates a single rotary valve of the type studied. Before this work, the valve was replaced every year on a fixed schedule, without any assessment of its condition or remaining service life. Vibration was recorded in three sessions at this valve position: on the service-exhausted unit immediately before its replacement, on the newly commissioned replacement unit, and on the same unit after three months of operation. The three recordings therefore cover the start, early operation and end of the valve's service life. The lifecycle stages are defined by the facility's maintenance schedule rather than by physical inspection or wear measurement; they denote time in service rather than a measured degree of wear, and differences between sessions may also reflect variations in operating conditions.

The sensor node was mounted horizontally on top of the rotor housing (Fig.~\ref{Real System}). With $N=1024$ samples per FFT frame and $\Delta f=3.342$ Hz (Section~\ref{sec:fft}), the effective sampling frequency is $f_s=N\Delta f\approx3422.5$ Hz, giving a Nyquist frequency of about 1711 Hz and a frame duration of about 0.30 s.
The spectra and the time-domain features were computed on the sensor node by STMicroelectronics' vibration-analysis firmware, used unmodified with its default settings. Before each FFT, the firmware removes the DC (gravity) component of the acceleration signal and multiplies each frame by a window function; the spectra of the overlapping frames acquired during each acquisition period are then averaged. No further filtering was applied.

The dataset used for the machine-learning models contains 1,287 records with no missing values, recorded on the service-exhausted and the newly commissioned valve. Each record contains the peak acceleration ($acc\_x$, $acc\_y$, $acc\_z$, in m/s$^2$) and the RMS value ($rms\_x$, $rms\_y$, $rms\_z$) of each axis reported by the sensor node; timestamps were removed. The composite magnitude $rms\_mag=\sqrt{rms\_x^2+rms\_y^2+rms\_z^2}$ was added (Section~\ref{sec:cleaning}). Table~\ref{tab:dataset} summarizes the features. The peak-acceleration channels reach the full-scale limit of the accelerometer (39.19 m/s$^2$, about 4 g) in 2.3--7.3\% of the records.

\begin{table}[h]
\centering
\caption{Descriptive statistics of the machine-learning dataset ($n=1287$ records; peak acceleration in m/s$^2$, RMS velocity in mm/s)}
\label{tab:dataset}
\small
\begin{tabular}{lrrrrr}
\toprule
\textbf{Feature} & \textbf{Mean} & \textbf{SD} & \textbf{Min} & \textbf{Median} & \textbf{Max} \\
\midrule
$acc\_x$ & 6.48 & 10.08 & 0.76 & 1.62 & 39.19 \\
$acc\_y$ & 10.11 & 11.65 & 0.77 & 2.73 & 39.19 \\
$acc\_z$ & 8.64 & 13.63 & 0.72 & 1.88 & 39.19 \\
$rms\_x$ & 0.159 & 0.129 & 0.080 & 0.145 & 2.901 \\
$rms\_y$ & 0.323 & 0.229 & 0.146 & 0.296 & 4.770 \\
$rms\_z$ & 0.410 & 0.255 & 0.222 & 0.384 & 4.522 \\
$rms\_mag$ & 0.550 & 0.360 & 0.315 & 0.508 & 6.537 \\
\bottomrule
\end{tabular}
\end{table}

No established criterion exists for the physical condition of a worn rotary valve (Section~\ref{sec2}). The binary reference label used to train and evaluate the supervised models was therefore generated by an engineering rule on the composite magnitude: records with $rms\_mag \le 0.5$ were labelled anomalous (class 1) and all others normal (class 0). The rationale is that a high magnitude across all axes reflects the continuous vibration of the running rotor, whereas a lower magnitude departs from this operating signature. The threshold lies close to the median of $rms\_mag$ (0.508), giving 578 anomalous (44.9\%) and 709 normal (55.1\%) records. The label is computed record by record with no fitted parameters, so label generation transfers no information between training and test data.

The models were evaluated according to their learning paradigm. K-Means and DBSCAN are unsupervised and were fitted without labels on Min--Max-scaled features; they are assessed with the silhouette score and by the records they flag. Isolation Forest was used to count anomalies in each lifecycle stage (Table~\ref{tab:Anomalies}). The model was fitted separately to the peak acceleration of each axis. Records with a negative decision-function value were counted as outliers, and records with a value between 0 and 0.04, which were inside the decision boundary, were counted as anomalies. No random seed was fixed. The SVM variants and the EBM are supervised and were trained and tested on a single random record-level split (80/20, random seed 42), giving 1,029 training records (572 normal, 457 anomalous) and 258 test records (137 normal, 121 anomalous); the same split was used for all supervised models, which were trained on unscaled features. SVM hyperparameters were selected by 5-fold cross-validated grid search fitted on the full dataset, so the test records contributed to SVM model selection; repeating the selection on the training records only changes the linear-SVM test accuracy from 99.61\% to 99.22\%. The EBM was used with default hyperparameters, and an XGBoost classifier trained on the same split was used only to compute SHAP values (Section~\ref{sec:xaimethod}). Because the split is at record level, records from the same recording session appear in both the training and the test set.

\subsection{Frequency Domain Vibration Analysis for condition monitoring}\label{sec:fft}
To monitor the health of the rotary valve in real-time for regular intervals we have carried out vibrational analysis on the valve. The goal is to detect early signs of wear, faults, or anomalies, allowing for timely maintenance and preventing unexpected failures. Rotary valves are critical components in industrial systems, where their reliable and safe operation is essential to ensure process continuity and prevent unexpected failures. Vibration-based condition monitoring using Fourier Transform analysis helps detect issues like unbalance, misalignment, and structural resonance \cite{b12cm}. The Fast Fourier Transform (FFT) is a mathematical tool that converts time-domain vibration data into the frequency domain. According to \cite{b13cm}, Time domain, FFT, and DWT are efficient vibration analysis techniques. FFT is widely used in steady-state conditions to detect failures like gearbox issues, mechanical unbalance, and broken rotor bars in rotating machinery. FFT allows for the identification of fault-related frequencies, making it easier to diagnose issues and monitor changes over time.

FFT was performed on vibrational data from the SensorTile node.
The accelerometer captures data at a high sampling rate ($f_s\approx3422.5$ Hz; Section~\ref{sec:dataset}), with the FFT algorithm providing a frequency resolution of 3.342 Hz. The next subsection outlines the steps used for FFT analysis in this work. The accelerometer collects vibration data along the three axes (X, Y, Z) at the given high sampling rate, and the time-domain raw data is also stored for further analysis.

\subsection*{Mathematical Background}
The time domain vibration data is transformed into frequency domain using the FFT algorithm. The FFT of a sequence \( x[n] \) of length \( N \) is given by:

\[
X[k] = \sum_{n=0}^{N-1} x[n] \cdot e^{-i \frac{2\pi}{N} kn}
\]

where:
\begin{itemize}
    \item \( X[k] \) is the complex amplitude at frequency bin \( k \).
    \item \( x[n] \) is the \( n \)-th time-domain sample of the vibration signal.
    \item \( N \) is the total number of samples (1024 in this case).
    \item \( k \) is the frequency bin index (\( k = 0, 1, 2, \dots, N-1 \)).
    \item \( i \) is the imaginary unit (\( i^2 = -1 \)).
\end{itemize}

The FFT output is a spectrum of amplitudes at different frequencies. The frequency resolution (\( \Delta f \)) of the FFT is determined by the sampling rate (\( f_s \)) and the number of samples (\( N \)):

\[
\Delta f = \frac{f_s}{N}
\]

In this study, the frequency resolution is:

\[
\Delta f = \SI{3.3422852}{\hertz}
\]

This means each frequency bin represents a range of \SI{3.3422852}{\hertz}, and the frequency corresponding to the \( k \)-th bin is given by \( f_k = k \cdot \Delta f \).

The Nyquist frequency of about 1711 Hz covers the 10--1200 Hz band analysed in Section~\ref{sec:results}. At the low rotational speeds typical of rotary valves, the rotor-related fundamentals (shaft and vane-pass frequencies) lie below or close to $\Delta f$ and are not resolved by a 1024-point frame. The configuration is therefore used to track band-level changes in vibration amplitude, to which drive-train components, electrical supply harmonics and impact-excited structural resonances contribute, rather than to identify individual fault frequencies or early-stage degradation of specific components, which would require longer frames, envelope analysis or order tracking.

Peaks in the frequency spectrum are analyzed to identify fault-related frequencies. Changes in amplitude at a certain frequency over time may suggest worsening conditions. The FFT results are tracked over time to monitor changes in vibration patterns. For instance, if the amplitude at a specific frequency increases steadily, it may indicate progressive wear or damage. Based on the FFT analysis, maintenance actions can be planned proactively. For instance, if a bearing fault frequency is detected, the bearing can be replaced during the next scheduled maintenance. If imbalance is detected, the equipment can be balanced to prevent further damage.

However, there are certain limitations of using FFT alone as stated in the literature \cite{b14cm}, such as not being effective for capturing transient features over time, we have opted for ML-based anomaly detection and predictive maintenance algorithms as shown in the next section.

\subsection{Data processing and ML Model Implementation for predictive maintenance}
Data pre-processing and feature engineering are crucial steps in preparing raw data for analysis and modeling. Raw datasets often contain missing values, irrelevant information, and inconsistencies that can significantly affect the performance of machine learning algorithms. Several Data-Preprocessing steps have been performed to reduce the overall computational complexity of the models, and develop highly accurate anomaly detection and predictive maintenance system.

\subsubsection{Correlation Analysis}
Correlation is used to examine the strength and direction of the relationship between two variables. The correlation coefficient serves as a quantitative measure of this relationship \cite{Corr1}. Feature extraction in knowledge and data engineering involves identifying and eliminating irrelevant or redundant information to enhance processing efficiency. Even if a machine learning algorithm inherently selects features or disregards the issue, performing feature extraction beforehand can be advantageous. By reducing data dimensionality, the hypothesis space shrinks, enabling faster algorithm performance. This process can sometimes enhance classification accuracy for future predictions, while in other cases, it results in a more compact, interpretable outcome \cite{Corr2}.

Large volumes of data are generated from a 3-axis accelerometer and therefore need to be reduced for efficient on-edge deployment. To address this, correlation analysis is performed to determine which of the three axes contributes most significantly to spotting anomalies; axes with a high correlation coefficient can be neglected. The correlation coefficient between two variables is given by Pearson's correlation formula:

\[
r = \frac{\sum (X_i - \bar{X})(Y_i - \bar{Y})}{\sqrt{\sum (X_i - \bar{X})^2 \cdot \sum (Y_i - \bar{Y})^2}}
\]

where \( X_i \) and \( Y_i \) are individual data points, and \( \bar{X} \), \( \bar{Y} \) are the respective means of the two datasets.

Correlation was used as a redundancy screen rather than as a measure of class-discriminative information, and no formal threshold was applied. In the machine-learning dataset, the peak accelerations along X and Z are strongly correlated ($r=0.93$), whereas Y is only weakly correlated with X ($r=0.34$) and Z ($r=0.29$), so X is largely redundant with Z; the same pattern was observed in all three lifecycle stages (Fig.~\ref{correlation}). The screening was exploratory and used all records. To avoid feature-selection bias, the X-axis features were retained as inputs to the EBM, whose explanations independently assign them low importance (Section~\ref{sec:xai}).

\subsubsection{Data Cleaning}\label{sec:cleaning}
By dropping unnecessary columns, we eliminate redundant or irrelevant features that do not contribute to the predictive power of the model, thereby reducing complexity and enhancing computational efficiency. This cleaning process ensures that the remaining data is relevant and meaningful, which can lead to more accurate and reliable outcomes. In this regard, the time-based features were dropped as the aim of this research was to focus directly on anomaly detection and not on time-series data for reporting.
Removing timestamps affects only the per-record classifiers. Progressive change is assessed across the lifecycle stages through anomaly counts, slew rates and spectra (Section~\ref{sec:results}). In deployment, per-record flags can be aggregated into a rolling anomaly rate, so that isolated flags indicate transient events and a sustained upward trend indicates progressive wear.

Root Mean Square (RMS) change in accelerometer values reports a sudden spike or an anomaly in the dataset. In order to normalize these values and account for the overall change in all three axis, a magnitude vector was computed on the said features.
Computing the magnitude of a vector, such as the RMS magnitude in this context, is an example of feature engineering that can provide significant insights. The magnitude condenses multiple dimensions (RMS values for all 3-axis) into a single representative value, capturing the overall intensity or variability of the data in a simplified form. This derived feature can reveal patterns and relationships that may not be apparent when examining the original features individually.
Because $rms\_mag$ also defines the reference label (Section~\ref{sec:dataset}), its contribution to model accuracy cannot be assessed independently in this study.

\subsubsection{Data Normalization \& Scaling}
Finally, it is important to normalize and scale the data to known ranges that directly impacts model's accuracy. In this regard, Min-max scaling based normalization technique was applied to scale the features between a range of 0, and 1. This is particularly important when dealing with machine learning algorithms that are sensitive to the scale of the input data, such as K-Means clustering or gradient descent optimization. Scaling ensures that all features contribute equally to the distance calculations and model training, preventing features with larger ranges from dominating the results. As a result, the algorithm can perform more effectively, leading to faster convergence and better overall performance.
Min--Max scaling was applied only to the inputs of the clustering models; the supervised models were trained on unscaled features, so no scaling statistics were computed from the test data.

\begin{table*}[ht!]
    \centering
    \caption{Machine Learning Algorithms used and their tuning configurations}
    \label{tab:ModelConfig}
    \small
    \begin{tabularx}{\textwidth}{@{}l l X X@{}}
        \toprule
        Algorithm & Parameter & Parameter Description & Configuration \\
        \midrule
        \textbf{DBSCAN} & \texttt{eps} & Max.\ distance between points to be neighbors & Grid search 0.1--1.0 (step 0.1); final 0.2 \\
                        & \texttt{min\_samples} & Min.\ points to form a cluster & Tuned per density (2--10); final 10 \\
        \midrule
        \textbf{K-Means} & \texttt{n\_clusters} & Number of clusters & Silhouette Score for optimal $k$; final $k=3$ \\
                         & \texttt{max\_iter} & Max.\ iterations for convergence & 300 \\
                         & \texttt{tol} & Convergence tolerance & 1e-4 \\
        \midrule
        \textbf{Isolation Forest} & \texttt{n\_estimators} & Number of trees & 50 (initial); 200 (final) \\
                                  & \texttt{max\_samples} & Samples per tree & auto \\
                                  & \texttt{contamination} & Expected outlier proportion & auto (initial); 0.04 (final) \\
                                  & \texttt{max\_features} & Features per tree & 3 \\
        \midrule
        \textbf{SVM} (C-SVC) & \texttt{C} & Regularization parameter & 0.1, 1, 10, 100 \\
                     & \texttt{kernel} & Kernel type & linear, poly, rbf, sigmoid \\
                     & \texttt{gamma} & Kernel coefficient & Grid search 1, 0.1, 0.01, 0.001 \\
                     & Selection & Hyperparameter selection & 5-fold cross-validated grid search \\
        \midrule
        \textbf{EBM} & -- & -- & InterpretML defaults; random seed 42 \\
        \textbf{XGBoost} & -- & -- & Defaults; used only to compute SHAP values \\
        \bottomrule
    \end{tabularx}
\end{table*}

\subsection{ML Model Parameters}
As previously mentioned, four machine learning methods including DBSCAN, K-Means, Isolation Forest (IF), and Support Vector Machines (SVM) were used, where each technique was chosen according to its appropriateness for particular scenarios and the characteristics of the dataset.
The sensor data in this study demonstrate a high degree of correlation, with normal and anomalous instances showing distinct and clear separation, which for the supervised models follows from the definition of the reference label (Section~\ref{sec:dataset}). This property enables precise tuning of models to identify anomalies in specific sensor readings, especially variations along the y-axis and z-axis. The configuration parameters for each technique are selected to enhance detection performance, as outlined in Table \ref{tab:ModelConfig}, with their functions discussed in the following sections.

\subsubsection{Clustering-Based Algorithms - DBSCAN \& K-Means}
DBSCAN is a density-based clustering algorithm that is particularly effective for detecting anomalies in datasets characterized by low-density regions. Due to the distinct differentiation between normal and anomalous instances in sensor data, DBSCAN can be optimized for spike detection by adjusting its critical parameters \cite{b13}. The $\epsilon$ parameter defines the radius of the neighborhood surrounding a data point and can be adjusted to capture local variations in sensor readings. A reduced $\epsilon$ value increases sensitivity to minor variations, making it suitable for identifying spikes along the y-axis and z-axis. Whereas, the $\textit{min\_samples}$ parameter, which defines the minimum number of points necessary to establish a dense region, may be increased to mitigate false positives in noisy sensor data. Adjusting these parameters allows DBSCAN to effectively identify anomalies associated with notable deviations in sensor readings, which is more likely to occur due to increased device vibrations in the faulty equipment.

On the other hand, K-Means, a centroid-based clustering algorithm, partitions sensor data into $\textit{k}$ clusters using distance metrics, identifying anomalies as points that are distant from their nearest centroid \cite{b14, b15}. The distinct separation in the data allows for the optimization of K-Means to identify spikes by selecting an appropriate number of clusters $\textit{n\_clusters}$. The optimal $\textit{k}$ can be determined using methods such as the elbow method or Silhouette score, ensuring that anomalies are not clustered with normal data. In this research, an optimization function was used to iteratively select the best performing $\textit{k}$ based on the silhouette scores.

\subsubsection{Outlier Detection using IF}
Although IF is typically applied to high-dimensional datasets containing relatively few anomalies \cite{b16,b17}, it was also evaluated because the extracted features provided clear separation between normal and anomalous operating conditions. In this regard, the careful adjustment of the $\textit{n\_estimators}$ parameter, which specifies the number of base estimators (trees) in the forest, was tuned for the detection of outliers. Furthermore, by adjusting the $\textit{max\_samples}$ parameter, the base estimator was able to effectively capture local spikes in sensor readings.

\subsubsection{SVM for anomaly detection}
Support Vector Machines (SVMs) were also investigated to differentiate between normal and anomalous instances. By selecting an appropriate kernel function and optimizing the model hyperparameters, SVMs can effectively learn a decision boundary that separates the two classes.
The C-support vector classifier (scikit-learn SVC) was used with $acc\_y$ and $rms\_mag$ as inputs; the regularization parameter $C$, the kernel and, for non-linear kernels, the kernel coefficient $\gamma$ were selected by grid search, where smaller values of $C$ give a wider, more noise-tolerant margin and larger values fit the training data more closely~\cite{b18}. Parameter selection always presents challenges where model sensitivity and accuracy are always traded for each unique data set. The selection and tuning parameters of these techniques are outlined in Table \ref{tab:ModelConfig}.

\subsection{Explainable AI}\label{sec:xaimethod}
Since real-time anomaly detection is crucial for designing predictive solutions in the industry, it is imperative to focus on all major fault points in the system. In this regard, resource-constrained devices can provide an automated, cost-effective, and safer solution to record and investigate machine behavior and transmit potential faults in near real-time for preventive measures. However, resource-constrained devices, especially those deployed in real-time environments, require optimum resource utilization for steady and consistent operation. XAI models help in this regard in three ways. First, they identify the major features that contribute to a model's predictions. Next, the impact of local and global parameters can further help fine-tune and optimize the model for application-specific requirements. Finally, XAI models provide a detailed analysis of a model's behaviour and the reasoning behind its decisions, which is critical for the confidence and trustworthiness of a cost-effective, pre-emptive maintenance system.

A significant advantage of the data-driven approach using trained ML models is its ability to map the underlying relationship between input variables and model outputs. In contrast to the time- and frequency-based techniques discussed above, data-driven approaches learn the patterns and relationships within the data directly, enabling a scalable and adaptive approach without the need for explicit programming, provided adequate data is available for training. In order to address the model interpretability and explainability requirements outlined above, a range of leading XAI methods, including Shapley Additive Explanations (SHAP)~\cite{rev_lundberg2017} and Microsoft's Explainable Boosting Machines (EBM)~\cite{rev_nori2019}, are used to build models that not only accurately detect anomalies (unwanted vibrations), but also enable deeper insights into the contributing variables that the model utilizes for real-time anomaly detection, thus enabling PdM. It is worth mentioning that the SHAP model requires excessive computational resources, whereas EBM provides a more general, rounded approach by offering both global and local explanations based on the features and their direct impact on the model.

\begin{equation}
\label{EBM}
    g(x) = \beta_{o} + \sum\limits_{i=1}^{n}{f_{i}(x_{i})}
\end{equation}

Since EBMs are built on Generalized Additive Models (GAMs) that are typically represented as sum of functions, the model interpretability becomes easier since each contributing feature can be directly interpreted. Equation ~\ref{EBM} elaborates this behavior, where $ f_{i}(x_{i})$ gives a shape function that describes how the model output is impacted by each feature $x_{i}$. EBMs use gradient boosting over bagged datasets.
The EBM is the model that is evaluated, explained and proposed for deployment. SHAP values were computed for an XGBoost classifier trained on the same features and data split, used only as a model-agnostic cross-check of the EBM explanations; XGBoost is not part of the model comparison.

\subsection{Resource Consumption parameters for On-Edge deployment}\label{sec:resource}
To understand the resource consumption of AI models, it is necessary to understand the time complexity of these models. K-Means clustering has a time complexity of \(O(n \cdot k \cdot d)\), where \(n\) is the data size, \(k\) is the number of clusters, and \(d\) is the feature dimensions \cite{kmtc}. DBSCAN scales as \(O(n^2)\) or \(O(n \log n)\), depending on implementation \cite{dbtc}. For Isolation Forest, each isolation tree has an average depth of \(O(\log \psi)\), giving a per-tree construction cost of \(O(\psi \log \psi)\) and a total training complexity of \(O(t \psi \log \psi)\) for \(t\) trees, where \(\psi\) is the subsample size \cite{iftc}. SVM training complexity depends on the solver: standard quadratic-programming-based training is \(O(n^3)\), large-scale linear (e.g., SGD-based) solvers reduce this to \(O(nd)\), and kernelized SVM (e.g., RBF or polynomial) increases complexity to \(O(n^2d)\), where \(n\) is the number of training samples and \(d\) is the number of features.

For the framework, training and hyperparameter optimization scale with the number of records: for kernel SVMs a tenfold increase in records raises training time by two to three orders of magnitude, and DBSCAN without a spatial index scales quadratically, whereas K-Means, Isolation Forest and the EBM scale approximately linearly. Inference, in contrast, operates on a fixed-length feature vector per record and is independent of the data volume and of the sampling frequency: an EBM prediction requires one lookup per term, a linear SVM $O(d)$ operations and a kernel SVM $O(n_{sv}d)$ for $n_{sv}$ support vectors. A higher sampling frequency or longer FFT frame increases only the feature-extraction cost, which grows as $O(N\log N)$ per axis. Retraining can therefore be scheduled periodically, while per-record inference remains lightweight on the edge device.

The time complexity of models is essential to understand the resource utilization of the models. The important resources considered in our work are - memory utilization, GPU utilization and energy consumption of the models.

\subsection*{Resource Utilization and Energy Consumption Measurement}
CPU, memory, and GPU utilization are each calculated as the proportion of time or capacity in active use relative to the total available (i.e., usage / total $\times$ 100).
Memory and GPU utilization are system-wide percentages of the total RAM and of GPU active time recorded during each run, and therefore include the operating-system baseline. The libraries used (scikit-learn, InterpretML, XGBoost) execute on the CPU, so the GPU values reflect system-level activity rather than model computation and are not used to rank the models. Each resource value covers one complete execution of a model script, including data loading, training, hyperparameter optimization where applicable, and evaluation; it is not a per-record inference measurement, so runtimes reflect the cost of training a model on the device rather than its inference latency.
Resource use was recorded on two platforms. On the laptop used for development, memory and GPU utilization were recorded with the Python libraries psutil and GPUtil, and energy with PyJoules. On the Jetson Nano, where GPUtil and PyJoules cannot read the GPU and energy counters, memory, GPU utilization and power were recorded with NVIDIA's tegrastats utility and the jtop monitor.

The total energy consumption of a model during execution, as measured by PyJoules, is given by:

\begin{equation}
    E_{\text{total}} = \sum_{d \in D} E_d
\end{equation}

\begin{equation}
    E_d = \sum_{c \in C} P_{d,c} \times T_{\text{sample}}
\end{equation}

where:
\begin{itemize}
    \item \( D \) is the set of monitored devices (e.g., CPU, GPU).
    \item \( C \) is the set of power domains within a device (e.g., cores, memory).
    \item \( E_d \) is the total energy consumption of device \( d \).
    \item \( P_{d,c} \) is the power consumption (in Watts) of component \( c \) within device \( d \).
    \item \( T_{\text{sample}} \) is the time interval between two power samples (in seconds).
\end{itemize}

Since PyJoules provides \( E_d \) directly in Joules, the total energy consumption in Joules (J) is:

\begin{equation}
    E_{\text{total}} = \sum_{d \in D} \sum_{c \in C} P_{d,c} \times T_{\text{sample}}
\end{equation}

On the Jetson Nano, tegrastats and jtop report the power drawn by the module (total input, CPU and GPU rails, in mW) from its on-board power monitor, and energy was obtained from these readings over the duration of each run. All energy values are reported in joules. All these resources are important to achieve a highly energy-efficient Edge-AI system, and also provide near-real-time condition monitoring capabilities on the edge.

\section{Results}\label{sec:results}
\subsection{Initial Data Analysis}
\subsubsection{Frequency domain vibrational analysis}
\begin{figure*}[hbt!]
\centering

\subfloat[]{\includegraphics[width=0.8\columnwidth]{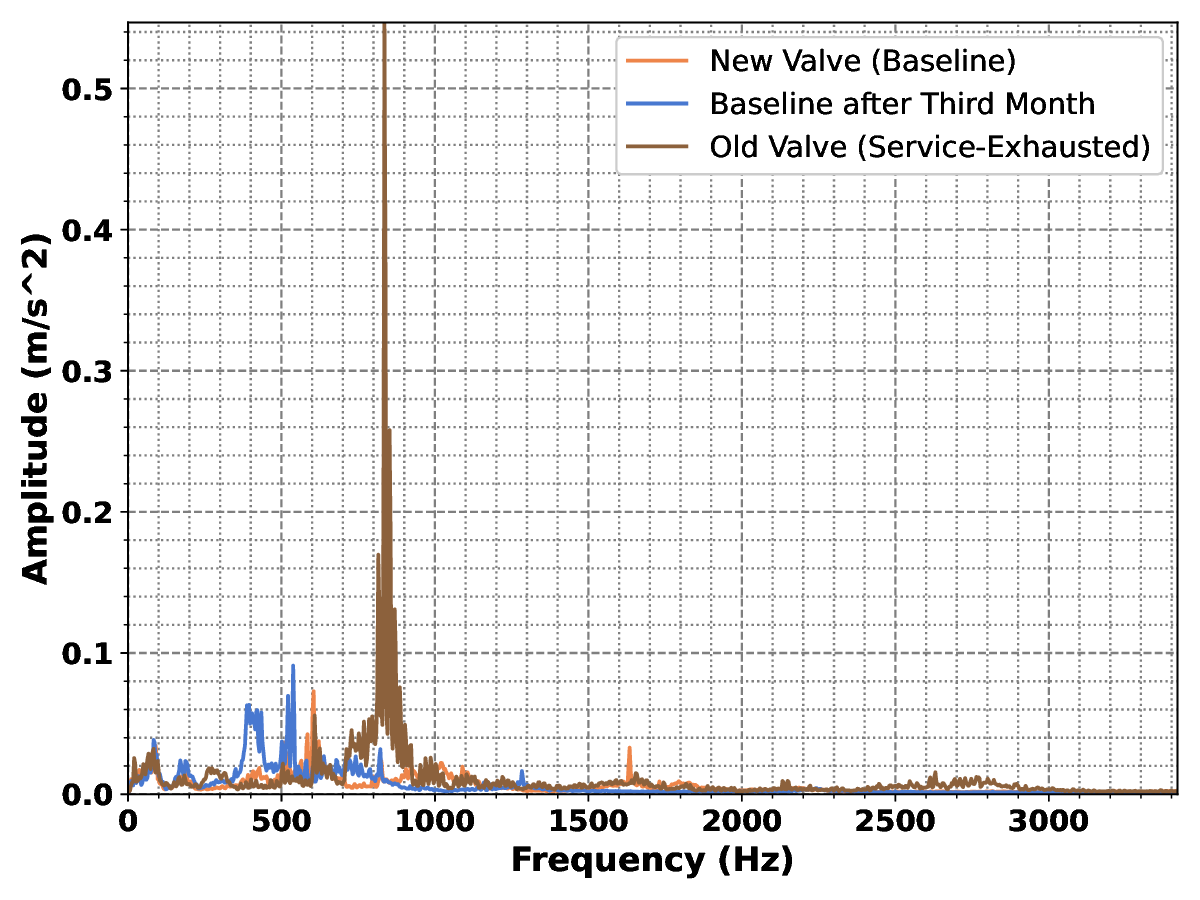}%
\label{fig:fft_x}}
\hfil
\subfloat[]{\includegraphics[width=0.8\columnwidth]{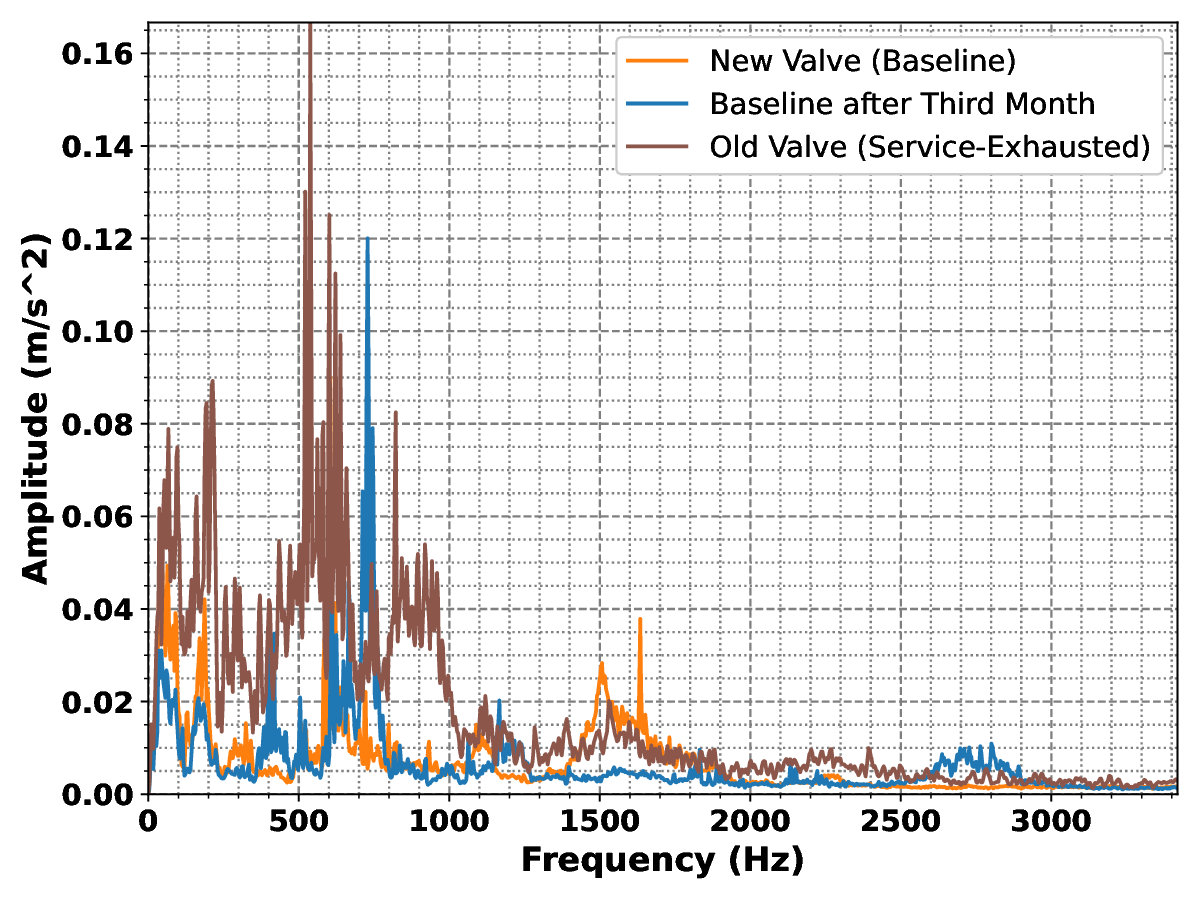}%
\label{fig:fft_y}}
\hfil
\subfloat[]{\includegraphics[width=0.8\columnwidth]{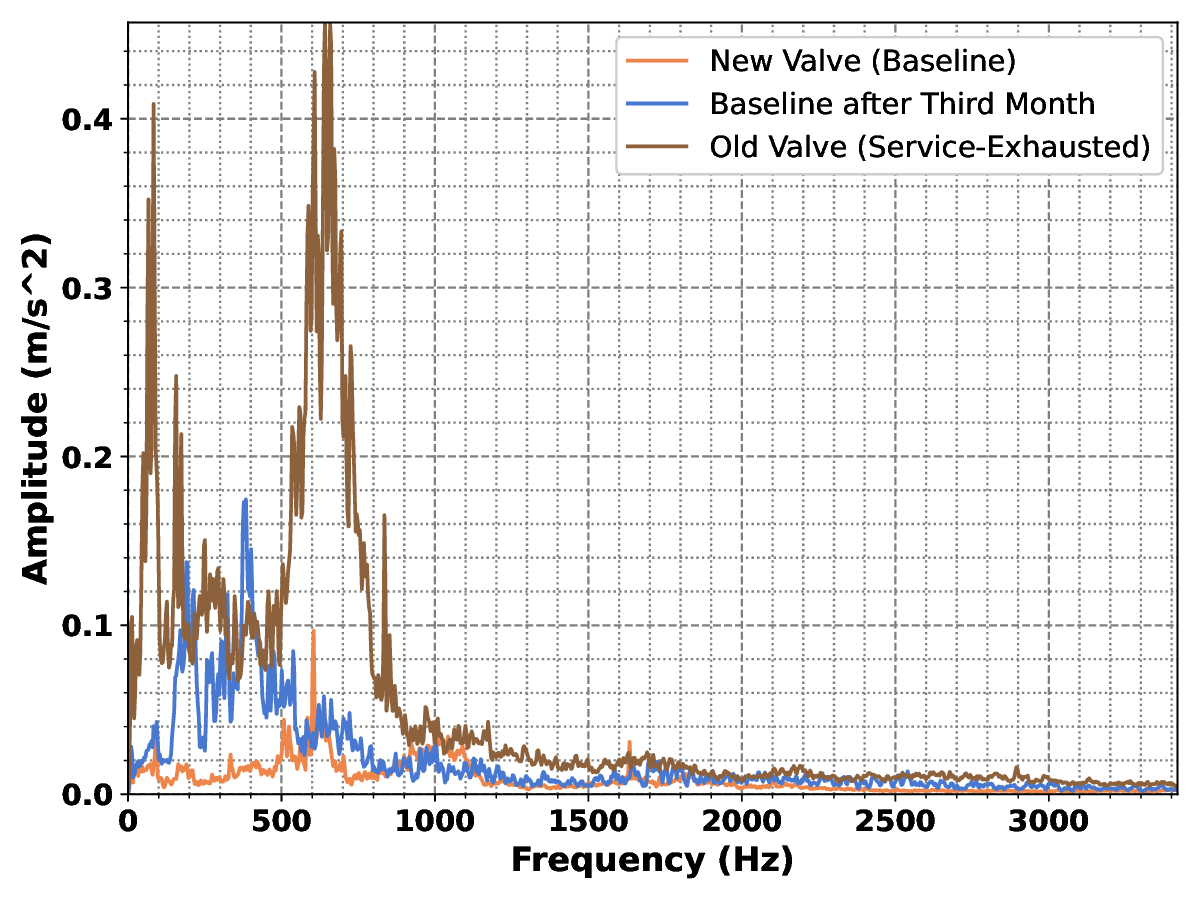}%
\label{fig:fft_z}}

\caption{Frequency vs Amplitude plots for vibrational data on (a) X-axis (b) Y-axis (c) Z-axis}
\label{Fig 4.}
\end{figure*}

Figure \ref{Fig 4.} shows the frequency domain plots for the vibrational data for X, Y, and Z axes respectively. The vibrational data was sampled at a rate sufficient to capture the relevant frequency range for the equipment under monitoring. The FFT frequency resolution was determined by the number of samples i.e. 1024 and the sampling rate, resulting in a resolution of 3.3422852 Hz.

The FFT amplitudes at specific frequencies were analyzed to identify patterns or deviations from baseline behavior, showing that the amplitudes rise for Y and Z axes between the frequency range of 10 Hz to 1200 Hz. It may correspond to mechanical resonances or faults. However, in terms of the newly commissioned valve (orange), the variation in amplitude is much lower compared to the old valve. There are a lot of variations in the amplitude for the old valve (brown), consistent with wear, while on the valve after three months (blue) there are some variations relative to the baseline. There is a sudden spike observed in the X axis plot (Fig.~\ref{fig:fft_x}) for the old valve. This indicates that the valve is producing more vibrations along all the three axis. These spectral changes are condition-dependent differences between recording sessions; they were not linked to specific components by inspection and may also reflect differences in operating conditions. Plausible sources in this band include impact-excited structural resonances, bearing defect frequencies, gear-mesh components of the drive and electrical components at 50 and 100 Hz~\cite{b12cm}.

\subsubsection{Correlation analysis}
\begin{figure}[hbt!]
\centering
\centerline{\includegraphics[width=\columnwidth]{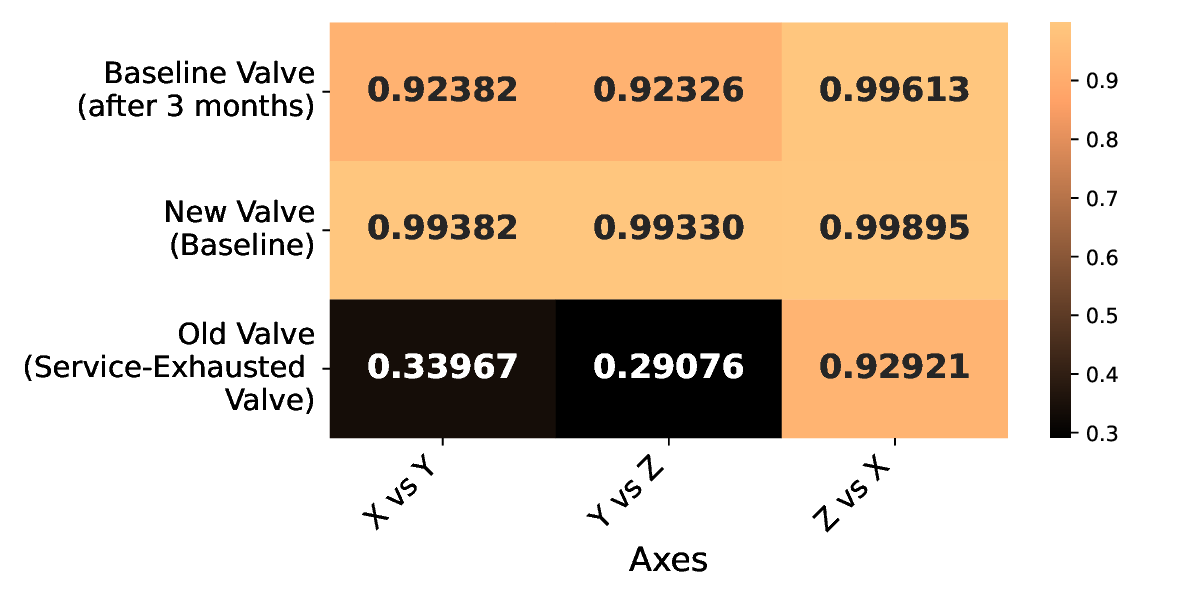}}
\caption{Correlation Heatmap of the 3-axis for the different months}
\label{correlation}
\end{figure}

Based on the correlation plots presented in Figure \ref{correlation}, it can be observed that the Y and Z axes exhibit the weakest correlation across all three data collection points. This observation can be explained by the horizontal placement of the board on the rotor, as shown in Figure \ref{Real System}. In this configuration, the sensor is subjected to more significant vibrations along the Y and Z axes. Consequently, anomalies in the Y and Z axes are prioritized for decision-making, with the Z-axis identified as the more critical axis for this purpose. Thus, only Y and Z axis were considered for Outlier detection, and providing real-time condition monitoring.

Additionally, the analysis reveals an interesting trend: as the rotor ages, the correlation between all three axes diminishes.
This suggests that the vibration becomes less coherent across axes as the valve ages, which may be associated with wear.

\subsection{Anomaly Detection }
 \subsubsection{Traditional Approach - Isolation Forest}
IF model was utilized to detect anomalies and outliers in the data collected from the rotary valve under different valve conditions: a service-exhausted valve, a new valve, and the same new valve after three months of operation. IF with 200 estimators and a contamination factor 0.04 (final configuration, Table~\ref{tab:ModelConfig}) was deployed on the edge device to detect anomalies in rotary valve vibrational data. The model calculated the anomaly score and classified data points as inliers or outliers, effectively identifying deviations in the dataset. The results are shown in Table \ref{tab:Anomalies}.

The results demonstrated a significant reduction in anomalies and outliers when transitioning from the service-exhausted valve to the new valve, consistent with the replacement. For the Y-axis and Z-axis datasets, the exhausted valve condition yielded 249 and 48 anomalies, respectively, compared to 6 and 8 anomalies for the new valve. After three months of operation on the new valve, anomalies slightly increased to 14 and 37 for the Y-axis and Z-axis. The outlier counts remained relatively stable across all conditions, with values ranging from 52 to 63. Slew rate was calculated, which provides a measure of signal variation. Slew rate for the exhausted valve was 1.9567 for Y-axis and 1.9422 for Z-axis. These values are relatively high, indicating rapid sample-to-sample changes in the vibration signal before replacement, which could suggest wear and tear. The values dropped significantly for the new valve, giving 0.2922 for Y-axis and 0.7647 for Z-axis, suggesting that the new valve operates more smoothly. After three months, the slew rate increased slightly (1.6574 for Y-axis and 1.0422 for Z-axis) but remained lower than the initial exhausted valve condition, indicating that the valve may be beginning to experience some wear as operating time accumulates.

\begin{table*}[hbt!]
\centering
\caption{Comparison of Anomalies, Outliers, and Slew Rate for Service Exhausted Valve, New Valve, and New Valve After 3 Months in Y-Axis and Z-Axis}
\label{tab:Anomalies}
\small
\resizebox{\textwidth}{!}{%
\begin{tabular}{lcccccc}
\toprule
 & \multicolumn{2}{c}{\textbf{Service Exhausted Valve}} & \multicolumn{2}{c}{\textbf{New Valve}} & \multicolumn{2}{c}{\textbf{New Valve (After 3 months)}} \\
\cmidrule(lr){2-3}\cmidrule(lr){4-5}\cmidrule(lr){6-7}
 & \textbf{Y-Axis} & \textbf{Z-Axis} & \textbf{Y-Axis} & \textbf{Z-Axis} & \textbf{Y-Axis} & \textbf{Z-Axis} \\
\midrule
\textbf{Anomalies} & 249 & 48 & \textbf{6} & \textbf{8} & 14 & 37 \\
\textbf{Outliers} & 63 & 52 & 60 & 58 & 52 & 52 \\
\textbf{Slew rate} & 1.9567 & 1.9422 & 0.2922 & 0.7647 & 1.6574 & 1.0422 \\
\bottomrule
\end{tabular}%
}
\end{table*}

 \subsubsection{Traditional Approach - K-Means, DBSCAN, SVM}
\begin{figure*}[hbt!]
\centering

\subfloat[Anomaly Detection using DBSCAN]{\includegraphics[width=0.99\columnwidth]{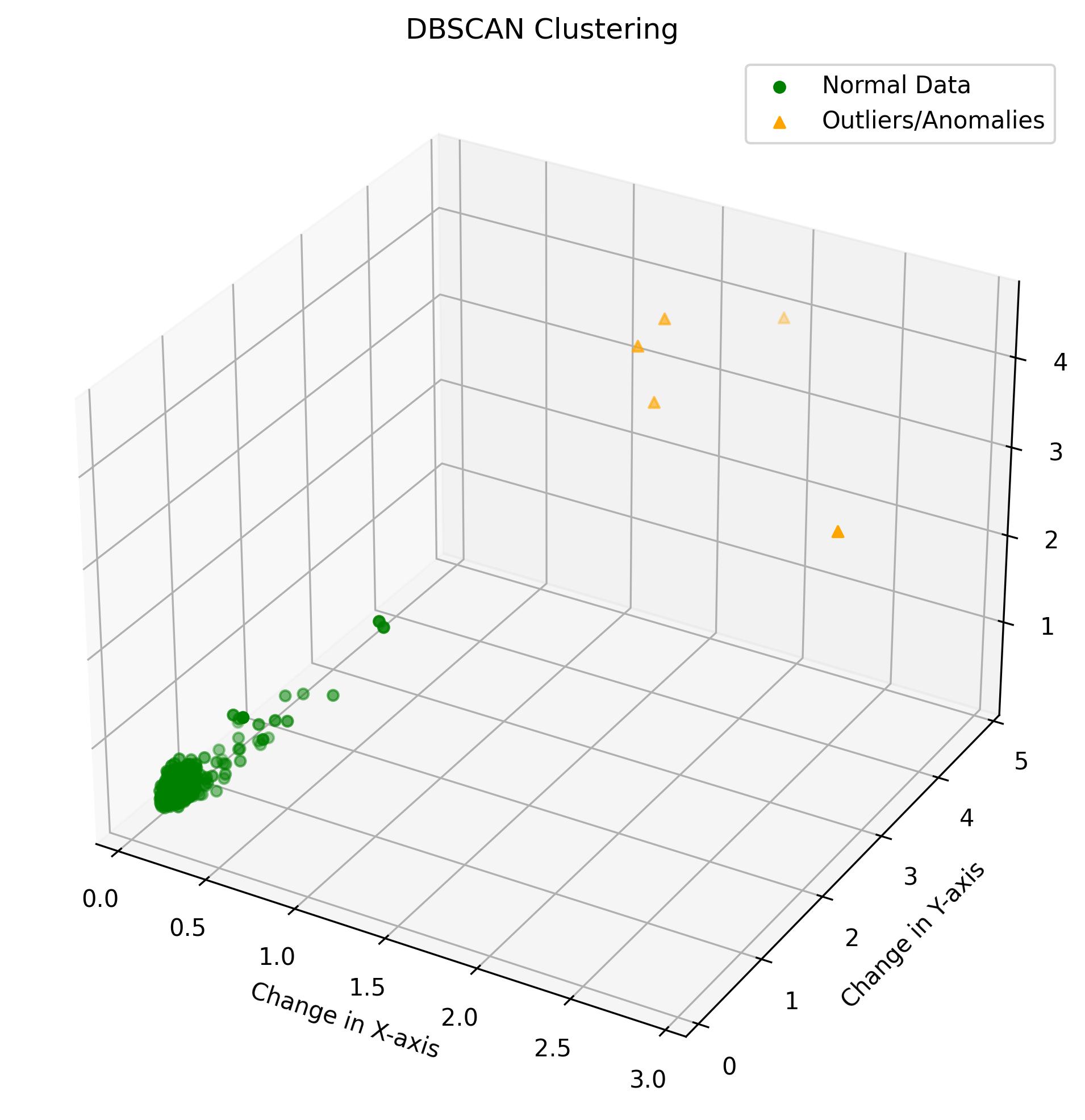}%
\label{fig:dbscan3d}}
\hfil
\subfloat[Anomaly Detection using K-Means]{\includegraphics[width=0.99\columnwidth]{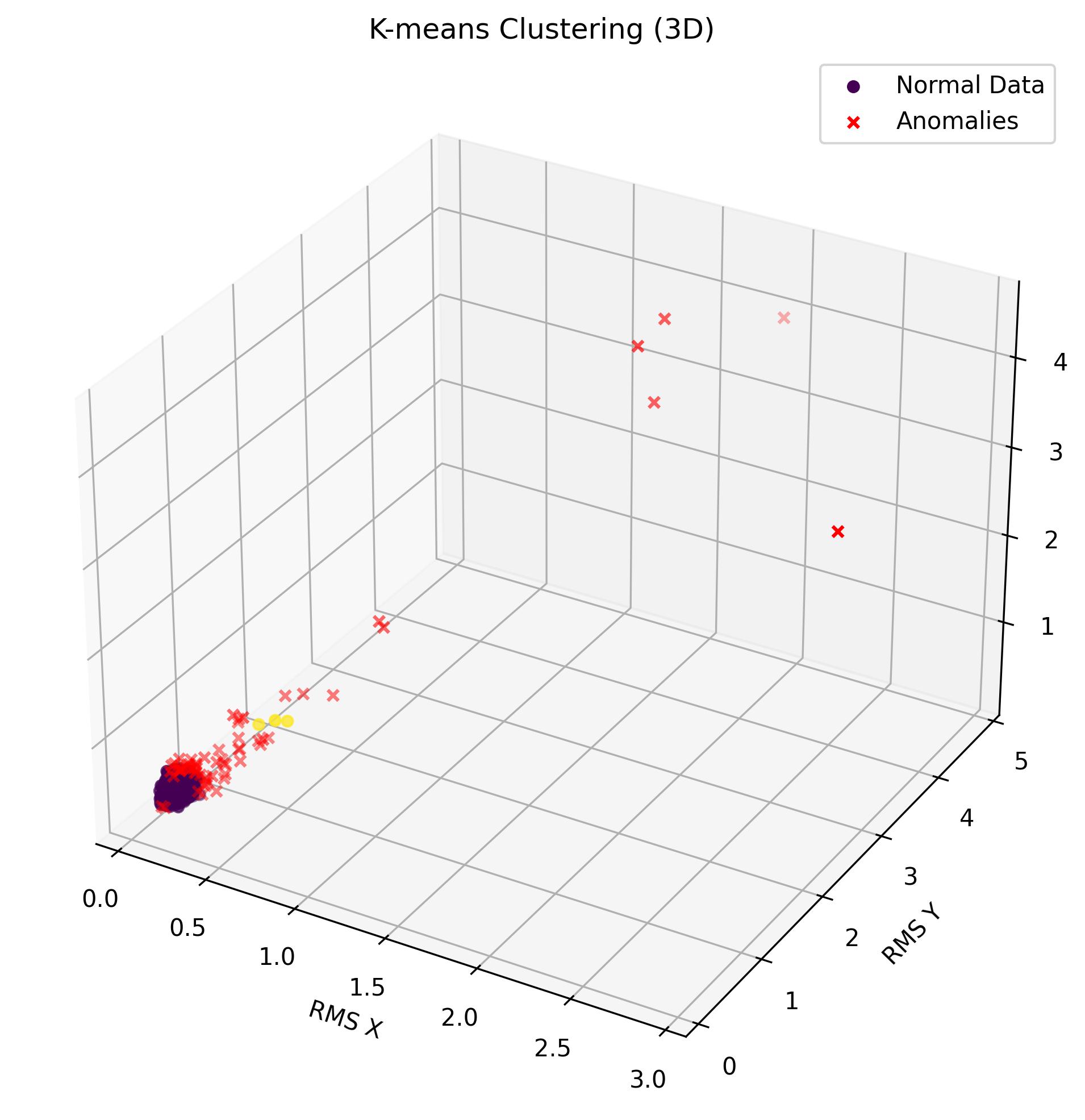}%
\label{fig:kmeans3d}}
\hfil

\caption{Clustering based techniques for Anomaly Detection on Target Variables}
\label{fig:3DPlots}
\end{figure*}

Figure \ref{fig:3DPlots} exhibits the output of the Density-Based Spatial Clustering of Applications with Noise (DBSCAN) algorithm applied to accelerometer data, emphasizing the differentiation between normal data points and outliers. The x-axis demonstrates a negligible change in the data, suggesting only slight variation in this dimension. The y-axis, representing the feature $\textit{acc\_y}$, records substantial vibrations generated by machine operations. The vibrations can be distinctly observed, indicating that the main dynamic variations occur along the y-axis. Additionally, the z-axis, represented by the feature $\textit{rms\_z}$, reflects the general rate of variation in the data, which is closely associated with the density of outliers. The outliers identified by DBSCAN are primarily affected by machine-induced vibrations, as demonstrated by the clustering pattern along the y and z axes.

In this context, the DBSCAN algorithm separates sparse outliers from dense regions of normal operation: with the final parameters it forms two clusters and labels 18 of the 1,287 records (1.4\%) as noise. The ability of the algorithm to discriminate between sparse regions of outliers and dense regions of normal data points is responsible for this behaviour. By concentrating on the most crucial features, $\textit{acc\_y}$ and $\textit{rms\_z}$, the model effectively captures the rate of change in the data, which is essential for detecting anomalies, generating unique density patterns that allow DBSCAN to effectively distinguish outliers. The algorithm is capable to manage noise and its non-parametric characteristics rendering it especially appropriate for this data type, where the distribution of normal and anomalous points may not adhere to a predetermined structure. Moreover, constraining the feature set to these two variables allows the model to attain expedited clustering and surface scans, thereby diminishing the computational load.

Similarly, the K-Means algorithm achieved good cluster separation (silhouette score 0.87) through its capacity to create distinct clusters based on the rate of change of the data using the same relevant feature set. The efficiency of K-Means can be notably compromised when multiple features simultaneously influence the input data stream. Vibration across various axes may cause cluster overlap, thus reducing the ability of the model to effectively detect anomalies. Despite this limitation, the K-Means algorithm demonstrates good separation and efficient resource utilization, positioning it as a viable option for real-time anomaly detection applications, contingent upon careful feature set selection and the simplicity of the data stream.

Because DBSCAN and K-Means are unsupervised, they were not scored against the reference label. With $k=3$, two small K-Means clusters contain the 25 records whose $rms\_mag$ exceeds 0.97, i.e., K-Means isolates rare high-energy events, and the 95th-percentile distance rule flags 65 records (5\%). Cluster membership is unrelated to the reference label (adjusted Rand index close to zero for both models): the clustering models flag rare high-amplitude events, whereas the reference label flags low-magnitude records, so the two capture different kinds of atypical behaviour.

\begin{figure}[hbt!]
\centering
\centerline{\includegraphics[width=\columnwidth]{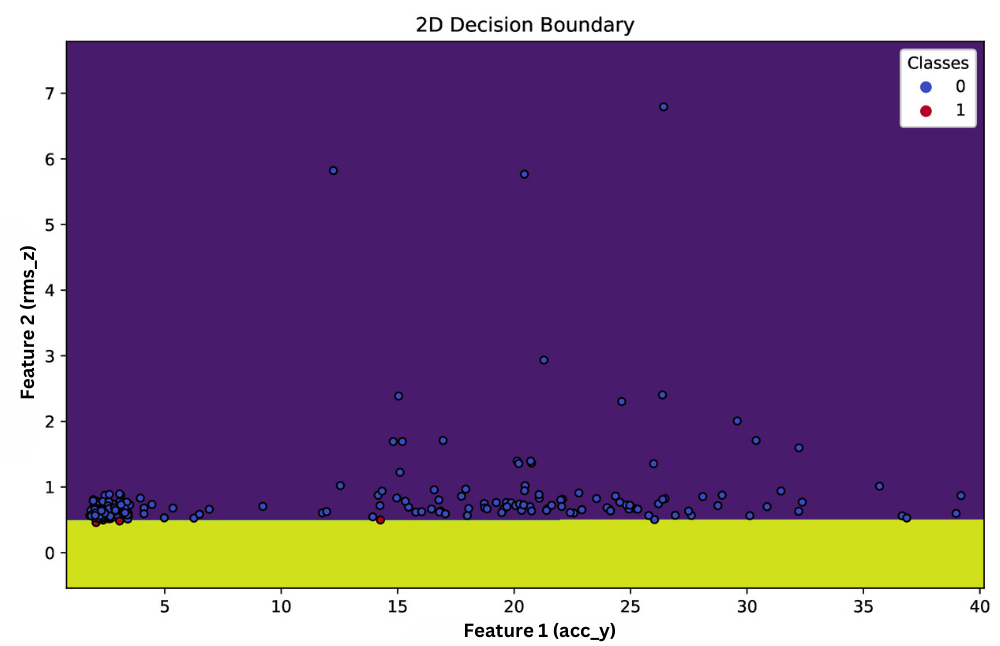}}
\caption{SVM feature boundary plot for target classes}
\label{fig:2D_Boundary}
\end{figure}

In addition, SVM were also used in the accelerometer data, resulting in an anomaly detection accuracy of 97.61\%. This performance demonstrates the capacity of the SVM model to accurately classify normal data points and anomalies through the establishment of optimal decision boundaries. The hyperplane of the model, depicted in Figure \ref{fig:2D_Boundary}, effectively isolates the characteristics by delineating their boundaries and distinctly differentiates anomalies from normal data. Because the reference label is a threshold on $rms\_mag$, one of the two SVM inputs, the class boundary is essentially a line at $rms\_mag\approx0.5$, which explains why a linear kernel suffices. This capability is especially important in situations that require accurate class separation, such as machine vibration monitoring. However, enhanced accuracy of SVM incurs a trade-off. The model necessitates considerably extended optimization and convergence durations compared to DBSCAN and K-Means. The prolonged runtime, along with elevated computational requirements, results in increased energy consumption, rendering SVM less appropriate for energy-limited devices or real-time applications where efficiency is critical. The performance of SVM is significantly influenced by the choice of kernel and the tuning of the model. The selection of kernel function (e.g., linear, radial basis function, or polynomial) and its associated parameters significantly impacts the model's generalization capability and classification accuracy. Furthermore, the presence of noise in the data can intensify these challenges, potentially reducing the performance of SVM and restricting its scalability in industrial applications.

Additionally, SVM requires precise tuning of application-specific parameters to achieve optimal outcome. The fine-tuning process is often time-consuming and complex, which limits the applicability of SVM as a general-purpose model for anomaly detection. SVM demonstrate superior accuracy in environments with ample computational resources. However, their drawbacks regarding energy efficiency, scalability, and sensitivity to noise and parameter selection hinder their practicality for extensive application in real-time, resource-constrained industrial settings.

\subsection{Classification metrics}\label{sec:metrics}
Table~\ref{tab:metrics} gives the confusion matrices and class-wise metrics of the supervised models on the 258 test records, with the anomalous class as positive. The EBM classifies all test records correctly, and the linear SVM misses one anomalous record (recall 0.992); neither raises a false alarm. The 95\% Clopper--Pearson confidence intervals of their accuracies overlap~\cite{rev_clopper1934}, so the two models are statistically indistinguishable on this test set. The XGBoost model used for SHAP also classifies all test records correctly, which confirms that the reference label is fully recoverable from the features (Section~\ref{sec:dataset}). The results come from a single hold-out split.

\begin{table*}[hbt!]
\centering
\caption{Test-set performance of the supervised models (258 records: 137 normal, 121 anomalous; positive class = anomalous). CI: 95\% Clopper--Pearson interval; Bal. acc.: balanced accuracy; MCC: Matthews correlation coefficient; FAR: false-alarm rate}
\label{tab:metrics}
\resizebox{\textwidth}{!}{%
\begin{tabular}{lccccccccccccc}
\toprule
\textbf{Model} & \textbf{TN} & \textbf{FP} & \textbf{FN} & \textbf{TP} & \textbf{Accuracy \% (95\% CI)} & \textbf{Precision} & \textbf{Recall} & \textbf{Specificity} & \textbf{F1} & \textbf{Bal. acc.} & \textbf{MCC} & \textbf{FAR} & \textbf{ROC-AUC} \\
\midrule
EBM            & 137 & 0  & 0  & 121 & 100.00 (98.58--100.00) & 1.000 & 1.000 & 1.000 & 1.000 & 1.000 & 1.000 & 0.000 & 1.000 \\
SVM (Linear)   & 137 & 0  & 1  & 120 & 99.61 (97.86--99.99)   & 1.000 & 0.992 & 1.000 & 0.996 & 0.996 & 0.992 & 0.000 & 1.000 \\
SVM (Sigmoid)  & 66  & 71 & 79 & 42  & 41.86 (35.77--48.14)   & 0.372 & 0.347 & 0.482 & 0.359 & 0.414 & $-0.172$ & 0.518 & 0.389 \\
XGBoost (SHAP only) & 137 & 0 & 0 & 121 & 100.00 (98.58--100.00) & 1.000 & 1.000 & 1.000 & 1.000 & 1.000 & 1.000 & 0.000 & 1.000 \\
\bottomrule
\end{tabular}%
}
\end{table*}

\subsection{Explainable insights using SHAP, EBM Boost}\label{sec:xai}

\begin{figure}[hbt!]
\centering
\centerline{\includegraphics[width=\columnwidth]{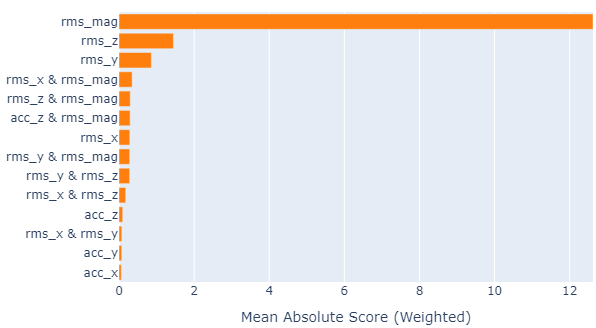}}
\caption{Importance ranking of dataset features.}
\label{fig:EBM_Global}
\end{figure}

\begin{figure*}[hbt!]
\centering

\subfloat[$rms\_mag$ feature density \& prediction score]{\includegraphics[width=0.99\columnwidth]{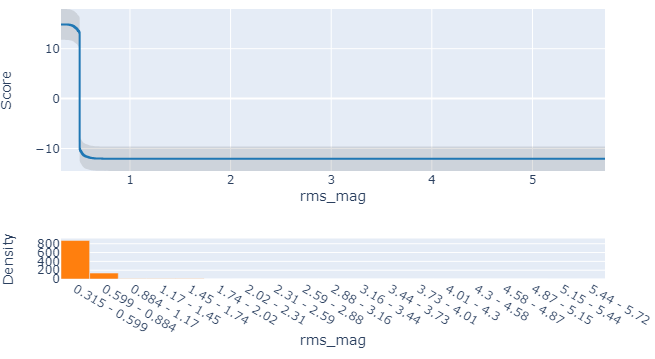}%
\label{rms_mag feature density and prediction score}}
\hfil
\subfloat[$rms\_z$ feature density \& prediction score]{\includegraphics[width=0.99\columnwidth]{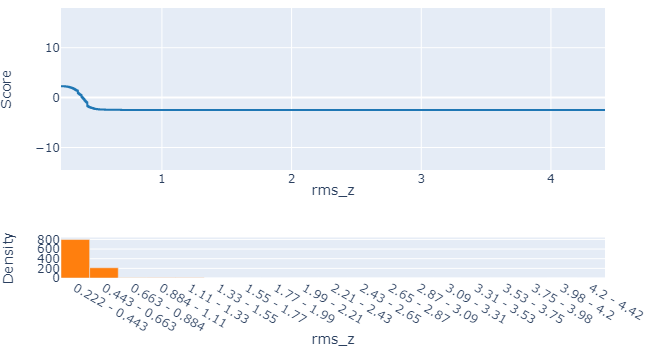}%
\label{rms_z feature density and prediction score}}
\hfil

\caption{EBM shape functions (term scores) and data density of the most significant features.}
\label{fig:EBM_Local}
\end{figure*}

Mean Absolute Scores for competing features enabled to identify the successful classified instances. In this regard, the accuracy score with a high degree of precision ensured that all instances were successfully classified and therefore interpreted by the models. This follows from the construction of the reference label, which is a threshold on $rms\_mag$, itself a model input (Section~\ref{sec:dataset}). The global interpretations in this regard are very useful in plotting the importance ranking of the most impactful parameters within the dataset, as given in Fig. \ref{fig:EBM_Global}, where $rms\_mag$, $rms\_z$ and $rms\_y$ (representing total rate of change at Z and Y-axis at a certain instance) directly impact the model output, while the linear change at X-axis ($rms\_x$) remains largely unimpactful, and other features and their combinations are least significant to the model's operation. The label is defined by $rms\_mag$, which therefore carries the decision. On average, $rms\_z$ contributes 54\% of $rms\_mag^2$, compared with 36\% for $rms\_y$ and 10\% for $rms\_x$, and these components are strongly correlated with $rms\_mag$ ($r=0.94$--$0.98$); boosting distributes part of the attribution among correlated inputs, so $rms\_z$ receives the largest share after $rms\_mag$. The dominance of $rms\_z$ thus reflects the mounting orientation of the sensor and is specific to this placement, although the inter-axis pattern was consistent across the three lifecycle stages (Fig.~\ref{correlation}).

\begin{figure}[hb!]
\centering
\centerline{\includegraphics[width=\columnwidth]{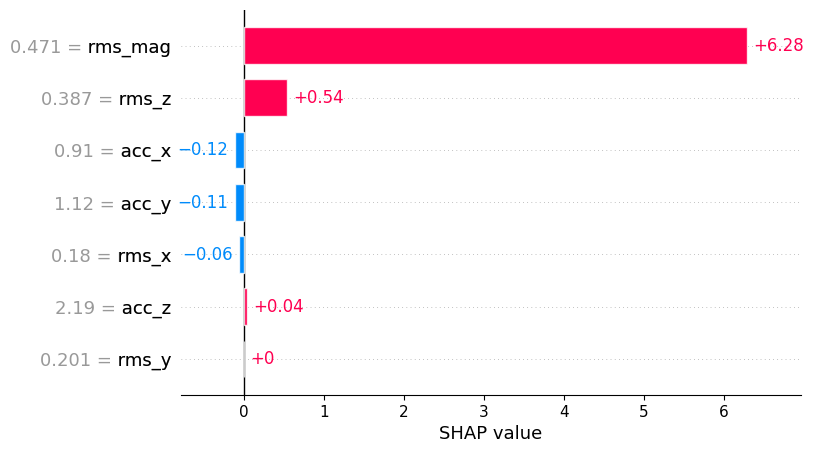}}
\caption{SHAP Interpretation of contributing features for one test record of the XGBoost model}
\label{fig:SHAP_features}
\end{figure}

Fig.~\ref{fig:SHAP_features} shows the SHAP decomposition for one test record of the XGBoost model. A negative SHAP value moves this prediction towards the normal class relative to the model's average output; it does not mean that the feature reduces model accuracy. For this record, $acc\_x$ and $acc\_y$ make small negative contributions. Whether removing a feature changes predictive performance can only be established by an ablation experiment.

\begin{figure*}[h!]
\centering

\subfloat[Model prediction for Normal Class]{\includegraphics[width=0.99\columnwidth]{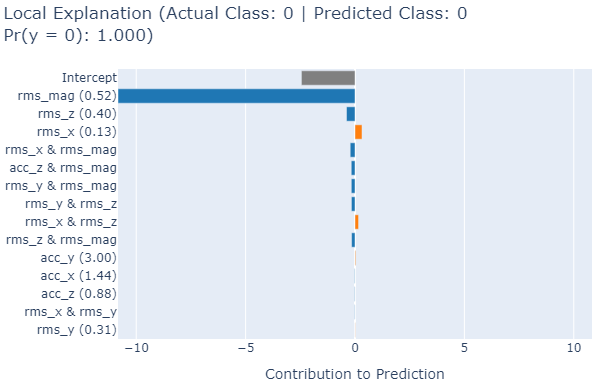}%
\label{fig:ebm_pred0}}
\hfil
\subfloat[Model prediction for Anomaly Class]{\includegraphics[width=0.99\columnwidth]{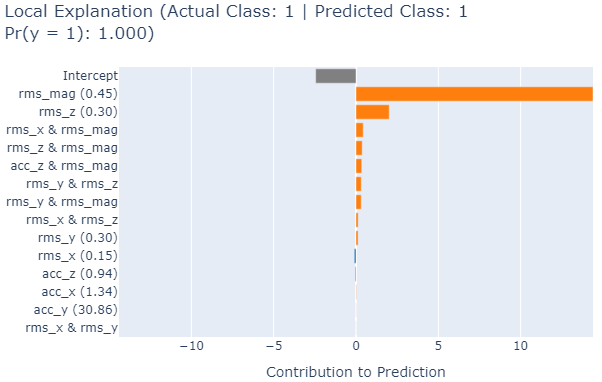}%
\label{fig:ebm_pred1}}
\hfil

\caption{EBM Local explanation for prediction.}
\label{fig:EBMPred}
\end{figure*}

This feature identification serves two critical purposes for real-time embedded ML integration in PdM settings.
First, identifying the most significant feature can indicate candidate reductions of the dataset, and thus the model training time, size, and complexity, making it suitable for low-resource, cost-effective embedded platforms that can be integrated with critical infrastructure; the actual effect of such a reduction must be confirmed by a controlled before-and-after experiment. Second, the consistency of this feature's impact on model performance enhances interpretability, which in turn builds trust and transparency in the model's operation.

EBM also provides a shape function for each feature. We discuss only the most significant features ($rms\_mag$ \& $rms\_z$) based on the above discussion, as given by Fig. \ref{fig:EBM_Local}. Each panel shows the score that each value of the feature adds to the prediction, together with the density of the feature. A positive score pushes the prediction towards the anomalous class and a negative score towards the normal class; the scores are not correlations. For $rms\_mag$ the score is positive below about 0.5 and negative above it: the even split between positive and negative scores is the signature of the labelling threshold and shows that $rms\_mag$ is the most decisive feature. For $rms\_z$, low values push the prediction towards the anomalous class and high values towards the normal class, consistent with $rms\_z$ being the dominant component of $rms\_mag$. The extent of a feature's positive or negative contribution signifies its importance, with greater absolute values denoting a more substantial effect on the prediction.

Next, the model predictability for each class (Actual vs. Predicted) was investigated. As illustrated in Fig. \ref{fig:EBMPred}, the model performs exceptionally well for both the normal and anomalous classes, due to the construction of the reference label noted earlier. For both classes, the prediction is driven by the RMS-magnitude terms ($rms\_mag$ and $rms\_z$), whereas the peak-acceleration features contribute little. Therefore, it is crucial to investigate model interpretability, enabling a 'glass-box' approach for traditional 'black-box' models, which improves not only model efficiency but also trustworthiness and transparency.

\subsection{Resource Consumption of AI models for real-time anomaly detection}

\subsubsection{Performance of cluster-based models}
DBSCAN is slightly computationally expensive, as expected from its time complexity (Section~\ref{sec:resource}). Table \ref{tab:kmeans_dbscan_resource} shows that DBSCAN consumes 65.1\% memory in comparison to 58.1\% for K-Means clustering. This clearly suggests that DBSCAN requires more storage space for clustering computations, due to its density-based approach which dynamically determines the number of clusters rather than assigning data points to fixed centroids as in K-Means clustering. Similarly, the GPU usage of DBSCAN is higher due to the nearest neighbor searches and region expansion operations, which require higher computational resources, leading to higher GPU utilization. The energy consumption of K-Means is relatively lower than that of DBSCAN, due to the higher memory and GPU utilization by DBSCAN leading to higher power consumption.
The values in Table~\ref{tab:kmeans_dbscan_resource} come from an individual clustering run with the final parameters, whereas those in Table~\ref{tab:model_comparison} cover the complete script, including the parameter search (36 optimization cycles for K-Means and 16 for DBSCAN), which accounts for their higher memory use (81.2\% and 86.1\%).

\begin{table}[h]
\centering
\caption{Resource Utilization: K-Means vs. DBSCAN (individual clustering run with the final parameters)}
\label{tab:kmeans_dbscan_resource}
\begin{tabular}{lcc}
\toprule
\textbf{Metric} & \textbf{K-Means} & \textbf{DBSCAN} \\
\midrule
Memory (\%) & 58.1 & 65.1 \\
GPU Utilization & Lower & Higher \\
Energy Consumption & Lower & Higher \\
\bottomrule
\end{tabular}
\end{table}

\subsubsection{Optimization performance of cluster-based models}
Before optimization, K-Means was clearly superior in performance with a silhouette score of 0.87 to 0.55 of DBSCAN, indicating that DBSCAN was struggling to form well-separated clusters. After optimization, both methods showed significant improvement, with DBSCAN catching up to K-Means, indicating that density-based clustering was successfully optimized for the dataset. K-Means took longer to optimize compared to DBSCAN; however, K-Means went through 36 optimization cycles as compared to 16 for DBSCAN, since K-Means updates cluster centroids iteratively, requiring more optimization cycles than DBSCAN's simpler optimization process. K-Means also has a lower optimization energy consumption than DBSCAN, making it more energy efficient, which is useful for low-power environments such as edge computing, IoT devices, and mobile devices. If energy efficiency is more important than optimization time, K-Means is the better choice for the application.

\begin{table}[h]
\centering
\caption{Optimization Behavior of K-Means and DBSCAN}
\label{tab:kmeans_dbscan_opt}
\small
\begin{tabularx}{\columnwidth}{@{}X c c@{}}
\toprule
\textbf{Metric} & \textbf{K-Means} & \textbf{DBSCAN} \\
\midrule
Silhouette score (pre-optimization) & 0.87 & 0.55 \\
Optimization cycles & 36 & 16 \\
Optimization time & Longer & Shorter \\
Optimization energy consumption & Lower & Higher \\
\bottomrule
\end{tabularx}
\end{table}

\subsubsection{Performance of SVM and IF}
To assess the computational demands of each model, we analyzed memory usage (\%), GPU usage (\%), energy consumption, and accuracy before and after optimization; results are consolidated in Table \ref{tab:model_comparison}. SVM (Poly) exhibits the highest memory and GPU utilization, making it the most computationally intensive model. In contrast, SVM (Sigmoid) shows the lowest GPU usage but consumes moderate memory. The Isolation Forest (IF) model, while exhibiting relatively moderate GPU usage, demonstrates high memory demands, suggesting increased dependency on memory resources for its optimization process. SVM with a linear kernel has shown balanced performance in terms of resource utilization, having the least memory utilization amongst the other models and also a moderate GPU usage, suggesting a lower computational burden.

Energy consumption varies significantly among models. SVM (Poly) exhibits the highest energy consumption (989 mJ), followed by IF (424 mJ), SVM (Linear) (369 mJ) and SVM (Sigmoid) (254 mJ). Although SVM (Sigmoid) consumes the least energy, its accuracy is unusable; among the models with acceptable accuracy, SVM (Linear) is the most energy-efficient.
These values come from individual runs of the final models, without the hyperparameter search, whereas the energy values in Table~\ref{tab:model_comparison} cover the complete script, including the search, and are therefore higher.

\subsubsection{Optimization performance of Accuracy-based models}
SVM (Linear) achieves an accuracy of 99.61\% after optimization, demonstrating that linear kernels can still be highly effective despite their simpler decision boundaries. SVM (RBF) also performs exceptionally well (98.81\%), demonstrating its effectiveness in capturing complex decision surfaces. SVM (Sigmoid), however, experiences a severe decline in accuracy (from 53.1\% to 41.86\%), indicating that optimization negatively impacts its performance. SVM (Poly) achieves 97.52\% accuracy post-optimization, suggesting that while computationally expensive, polynomial kernels still provide strong classification capabilities. The results suggest that SVM (Linear) remains a competitive choice, achieving high accuracy while requiring lower computational resources than SVM (Poly).

Understanding the relationship between optimization time and model execution time is essential for assessing the practical viability of each model. Table \ref{tab:svm_opt_time} presents optimization time (in seconds) for the SVM variants and IF. SVM (Poly) took the longest optimization time and also had the longest model runtime, making it computationally expensive. SVM (Linear) has a moderate optimization time, indicating that while the linear kernel benefits from optimization, its impact on execution time is not as significant as other SVM variants. SVM (Sigmoid) has the shortest optimization time, making it the fastest model amongst all; however, the accuracy drop noted above makes it irrelevant for our application. IF requires a longer optimization time but still achieves a reasonable runtime, comparable to SVM (Poly). These results highlight the trade-offs between kernel selection and optimization efficiency, with SVM (Linear) offering a balanced approach but still being outperformed by SVM (RBF) in runtime.

\begin{table}[h]
\centering
\caption{Optimization Time for SVM Variants and IF}
\label{tab:svm_opt_time}
\begin{tabular}{lc}
\toprule
\textbf{Model} & \textbf{Optimization Time (s)} \\
\midrule
SVM (Linear)  & $\sim$8.3 \\
SVM (Sigmoid) & 2 \\
SVM (Poly)    & 90 \\
IF            & 75 \\
\bottomrule
\end{tabular}
\end{table}

\subsubsection{Performance of XAI Model}
\begin{table*}[hbt!]
\centering
\caption{Accuracy and resource utilization on the edge device. (a) Unsupervised clustering models; (b) the Isolation Forest, which is unsupervised and has no label-based accuracy, and the models evaluated against the reference label; (c) reduction achieved by the EBM relative to each model in (b), computed as (model - EBM)/model; positive values mean that the EBM uses less. n/a: not measured or not applicable}
\label{tab:model_comparison}
\small
\resizebox{\textwidth}{!}{%
\begin{tabular}{lccccc}
\toprule
\multicolumn{6}{l}{\textbf{(a) Unsupervised clustering models}} \\
\midrule
\textbf{Model} & \textbf{Silhouette} & \textbf{Memory (\%)} & \textbf{GPU (\%)} & \textbf{Energy} & \textbf{Runtime (s)} \\
\midrule
K-Means & 0.87 & 81.2 & 3.01 & 92.42 & n/a \\
DBSCAN  & 0.55 & 86.1 & 7.86 & 100.1 & n/a \\
\midrule
\multicolumn{6}{l}{\textbf{(b) Isolation Forest and models evaluated against the reference label}} \\
\midrule
\textbf{Model} & \textbf{Accuracy (\%)} & \textbf{Memory (\%)} & \textbf{GPU (\%)} & \textbf{Energy (J)} & \textbf{Runtime (s)} \\
\midrule
IF             & n/a    & 85.1 & 33 & 2.13 & 213.4 \\
SVM (Linear)   & 99.61  & 76.8 & 54 & 2.10 & 114.9 \\
SVM (RBF)      & 98.81  & 82.7 & 68 & 2.81 & 108.7 \\
SVM (Sigmoid)  & 41.86  & 89.0 & 41 & 2.85 & 107.7 \\
SVM (Poly)     & 97.52  & 94.0 & 59 & 3.89 & 216.6 \\
EBM            & 100.00 & 55.0 & 11 & 1.25 & 30.02 \\
\midrule
\multicolumn{6}{l}{\textbf{(c) Reduction by the EBM relative to each model in (b)}} \\
\midrule
\textbf{Model} & \textbf{Accuracy gain (pp)} & \textbf{Memory (\%)} & \textbf{GPU (\%)} & \textbf{Energy (\%)} & \textbf{Runtime (\%)} \\
\midrule
IF             & n/a    & 35.4 & 66.7 & 41.3 & 85.9 \\
SVM (Linear)   & +0.39  & 28.4 & 79.6 & 40.5 & 73.9 \\
SVM (RBF)      & +1.19  & 33.5 & 83.8 & 55.5 & 72.4 \\
SVM (Sigmoid)  & +58.14 & 38.2 & 73.2 & 56.1 & 72.1 \\
SVM (Poly)     & +2.48  & 41.5 & 81.4 & 67.9 & 86.1 \\
\bottomrule
\end{tabular}%
}
\end{table*}

Table~\ref{tab:model_comparison} separates the unsupervised clustering models from the models evaluated against the reference label and compares resource use within each group. Among the label-evaluated models, the EBM has the highest accuracy (statistically indistinguishable from SVM (Linear); Section~\ref{sec:metrics}), the lowest memory utilization (55\%), the lowest energy consumption (1.25 J) and the shortest runtime (30.02 s), reducing memory by 28.4--41.5\%, energy by 40.5--67.9\% and runtime by 72.1--86.1\% relative to the other models in this group. Its GPU utilization (11\%) is the lowest in this group but higher than that of K-Means (3.01\%) and DBSCAN (7.86\%); because none of the models runs on the GPU, these values reflect system-level activity.

The EBM uses less memory than the clustering models (55\% versus 81.2\% and 86.1\%); their runtimes and label-based accuracies were not measured, so no improvement is computed for them. The large runtime and energy reductions also reflect that the EBM was used with default hyperparameters, whereas the SVMs and Isolation Forest included a hyperparameter search.

\subsection{Discussion}
Although neural networks and ML based algorithms can provide a simple yet scalable approach for anomaly detection, the application and resource constraints must also be analyzed for industrial systems. For instance, real-time on-device data acquisition, detection and reporting are some of the fundamental requirements in industrial environments. Due to the complex architecture, limited space and storage, strict resource utilization limitations are inherently imposed. This further dictates the use of additional or external resources, such as data network for inferencing, storage network for data off-loading. Therefore, a cost-effective integrated solution is required that can run pre-trained lightweight ML models for inferencing and anomaly detection. This necessitates decoupling the ML model complexity by aligning the metrics with the resource requirements. Explainable Artificial Intelligence (XAI) may play a crucial role in enhancing time frequency analysis (TFA) and neural networks for anomaly detection by providing transparency and interpretability to complex AI models. By making AI processes more transparent and understandable, XAI not only enhances trust and accountability in AI systems but also enables experts to collaborate effectively with intelligent systems, leading to improved decision-making processes and model reliability. Furthermore, XAI may enable the design of less complex, small-footprint lightweight ML models that could be deployed in any industrial environment. In this regard, data pre-processing, feature engineering and XAI inferencing is required to select the optimal ML model for sensor data classification.

\textbf{From condition monitoring towards predictive maintenance:} The results show that vibration features differ between lifecycle stages of the monitored valve and that lightweight models can flag atypical records on an edge device and explain their decisions; they do not constitute failure prediction or RUL estimation. The framework provides the condition-monitoring layer on which PdM can be built in three steps: (i) aggregating per-record anomaly flags into a health indicator, for which the Isolation Forest anomaly counts already give a first indication of an increase over time (Z-axis: 8 for the new valve, 37 after three months and 48 for the service-exhausted valve; Table~\ref{tab:Anomalies}); (ii) calibrating an end-of-life threshold from the health indicator of the service-exhausted valve; and (iii) fitting a degradation model to run-to-replacement data to forecast the time to that threshold, i.e., the RUL. At the pilot facility, the condition information from the system already supported keeping the valve in service for a few months beyond its scheduled annual replacement, a first step from fixed-interval towards condition-based replacement.

\textbf{Practical deployment:} The implemented components are the sensor node, the Bluetooth link to a mobile application and to the edge device, and on-device model training, inference and explanation. To limit false alarms, a maintenance alert should be raised only when the rolling anomaly rate stays above its commissioning baseline over a confirmation window. Each alert should be shown with its EBM explanation so that an operator can confirm it before intervening, until the system is validated, the existing replacement schedule remains the fail-safe. Because data are processed on premises, raw process data need not leave the site. A deployment should use encrypted and authenticated bluetooth pairing, access control on the edge device and signed model updates to protect against manipulated sensor data or models.

\textbf{Limitations and transferability:} The study covers a single valve at one pilot-scale facility, where it is the only valve of this type, and data-governance restrictions prevented data collection from other equipment in the plant. The records are therefore not independent equipment-level observations, and the near-perfect supervised scores should not be extrapolated to other valves, operating conditions or plants. The processing pipeline (tri-axial vibration, spectral and RMS features, redundancy screening, an interpretable model and on-device resource accounting) can be applied to other rotating equipment such as pumps, motors and conveyors, but the reference threshold, dominant axis, feature ranking and model parameters must be re-established for each machine and mounting position. Further limitations are the record-level random split, the absence of a feature ablation (including a controlled test of XAI-guided feature reduction), single-run resource measurements without hardware power validation or per-window latency measurement, and the use of a single sensing modality. Because the reference label is a single threshold, simpler interpretable models such as a shallow decision tree would also reproduce it; the case for the EBM rests on its resource footprint and explanations rather than on accuracy.

\section{Conclusion}
This study presented an Edge-AI framework for vibration-based condition monitoring of a rotary valve in a dairy pilot plant, combining frequency-domain analysis, machine learning and explainable AI on an edge device, as a step towards predictive maintenance. Spectra and inter-axis correlations differed between a service-exhausted valve, a newly commissioned valve and the same valve after three months of operation, and Isolation Forest anomaly counts rose from the new to the service-exhausted condition. Against a rule-based reference label derived from the RMS vibration magnitude, the supervised models reproduced the label with high fidelity (EBM 100\%, linear SVM 99.61\%), and among them the EBM required the least memory, energy and runtime on the edge device. Its explanations recovered the labelling threshold and showed that the Z-axis component dominates the vibration magnitude at this sensor placement, illustrating how a glass-box model can explain its decisions and point to candidate feature reductions for constrained devices. At the pilot facility, this condition information supported keeping the valve in service beyond its fixed annual replacement.

The results come from a single installation with rule-based labels and do not yet constitute failure prediction. Future work will (i) validate the approach on further valves and sites with inspection-confirmed labels and equipment-independent testing, (ii) build a health indicator from the anomaly rate and estimate RUL, (iii) perform feature ablations, including a controlled test of XAI-guided feature reduction, (iv) measure per-window latency and hardware-validated energy under continuous operation, and (v) benchmark other interpretable and lightweight deep-learning models.

\backmatter

\bmhead{Acknowledgements}
The authors would like to thank Moorepark Technology Limited Fermoy, Co.Cork, Ireland) for the use of pilot scale facilities,  providing data to be collected for experimentation of Use Case - condition monitoring of rotary valves for the demonstration of technology and proof of concept.

\bmhead{Funding}
This research was supported by the Science Foundation Ireland and the Department of Agriculture, Food and Marine on behalf of the Government of Ireland Vistamilk Research Center under grant 16/RC/3835.

\bmhead{Ethics Declaration}
\bmhead{Conflict of Interest}
The authors declare that they have no competing interests.
\bmhead{Data availability }
The datasets generated and/or analysed during the current study are not publicly available due to confidentiality restrictions associated with data collected from the pilot study site. However, a limited, de-identified sample dataset may be made available by the corresponding author upon reasonable request, subject to applicable confidentiality and data-sharing restrictions. The characteristics of the dataset are summarized in Section~\ref{sec:dataset} and Table~\ref{tab:dataset}.

\bibliography{sn-bibliography}%

\end{document}